\documentclass{amsart}

\usepackage{color}  
\usepackage{centernot} 
\usepackage{graphicx} 
\usepackage{epstopdf} 
\usepackage{amsaddr} 

\usepackage[numbers, sort&compress]{natbib}
\makeatletter 
\renewcommand\@biblabel[1]{#1.} 
\makeatother %

\usepackage{booktabs}
\usepackage{tabularx}
\usepackage{colortbl}
\usepackage[table]{xcolor}
\usepackage{multirow}
\usepackage{supertabular}
\DeclareMathOperator*{\argmax}{arg\,max}

\newcommand{\ie} {\textit{i.e.}}
\newcommand{\dps} {\displaystyle}
\newcommand{\ud}{\mathrm{d}}
\renewcommand{\Re}{\mathbb{R}}

\newcommand{\boma}[1]{\mbox{\boldmath ${#1}$}}

\newtheorem{exemplo}{Example}
\newtheorem{theorem}{Theorem}
\newtheorem{definition}{Definition}

\definecolor{lightblue}{rgb}{0.8,0.85,1}

\newcommand\BibTeX{{\rmfamily B\kern-.05em \textsc{i\kern-.025em b}\kern-.08em
T\kern-.1667em\lower.7ex\hbox{E}\kern-.125emX}}

\begin{document}

\title[Theoretical Evaluation of Feature Selection Methods based on MI]{Theoretical Evaluation of Feature Selection Methods\\ based on Mutual Information}

\author{Cl\'{a}udia Pascoal$^1$, M. Ros\'{a}rio Oliveira$^1$, Ant\'{o}nio Pacheco$^1$ and Rui Valadas$^2$}

\address{$^1$ CEMAT and Dep. Mathematics,\\
       Instituto Superior T{\'e}cnico, Universidade de Lisboa,\\
       Av. Rovisco Pais, 1049-001 Lisboa, Portugal.\\ $^2$ IT and Dep. Electrical and Computer Engineering,\\
              Instituto Superior T{\'e}cnico, Universidade de Lisboa,\\
              Av. Rovisco Pais, 1049-001 Lisboa, Portugal.}

\begin{abstract}
Feature selection methods are usually evaluated by wrapping specific classifiers and datasets in the evaluation process, resulting very often in unfair comparisons between methods. In this work, we develop a theoretical framework that allows obtaining the true feature ordering of two-dimensional sequential forward feature selection methods based on mutual information, which is independent of entropy or mutual information estimation methods, classifiers, or datasets, and leads to an undoubtful comparison of the methods. Moreover, the theoretical framework unveils problems intrinsic to some methods that are otherwise difficult to detect, namely inconsistencies in the construction of the objective function used to select the candidate features, due to various types of indeterminations and to the possibility of the entropy of continuous random variables taking null and negative values.
\end{abstract}


\maketitle

\section{Introduction}

Feature selection is an important 
step in statistical learning problems involving high-dimensional data, e.g. regression and classification. Its main benefits are to facilitate data visualization and data understanding, reduce measurement and storage requirements, reduce training and utilization times, and improve predictor performance~\cite{Guyon:2003}. Due to its importance, many excellent surveys on feature selection methods have been produced over the years \cite{Ferri:1994, Dash:1997, Jain:1997, Kudo:2000, Saeys.et.al:2007, Brown:2012, Vergara.Estevez:2014, Bennasar.Setchi:2015, Freeman.et.al:2015, Vinh.et.al:2016}.

The main goal of feature selection is to find a subset of features that leads to optimal performance of the learning process. This involves keeping relevant features, and removing those that are irrelevant or redundant~\cite{Vergara.Estevez:2014, Vinh.et.al:2016, John.Kohavi.Pfleger:1994, Koller.Sahami:1996, Blum:1997, Yu.Liu:2004}.

Feature selection methods can be classified into three  categories~\cite{Saeys.et.al:2007,Bennasar.Setchi:2015}: \textit{filters} methods, \textit{wrappers} methods, and  \textit{embedded} methods. Wrapper methods embrace the classifier in the selection process; the features are selected according to classifier performance metrics such as recall and precision~\cite{Saeys.et.al:2007, Kohavi:1997, Sindhwani:2004}. Filter methods select the features independently of the classifier; the selection process tries to find the subset of features that is most associated with the class variable~\cite{ Saeys.et.al:2007,Vergara.Estevez:2014,Bennasar.Setchi:2015}. Embedded methods combine the filter selection stage with the learning step~\cite{Saeys.et.al:2007,Vergara.Estevez:2014,Lal.et.al:2006}.

Wrapper methods have two relevant disadvantages: their large computational complexity and their dependence on a specific classifier.  Apart from being classifier independent, filter methods are computationally less demanding than wrapper methods and, as a result, are more suitable for high-dimensional problems.   Embedded methods are also classifier dependent, but less onerous in computational complexity and less sensitive to over-fitting than wrapper methods. However,  embedded methods are designed specifically to a certain classifier, which constrains their generalization~\cite{Saeys.et.al:2007,Bennasar.Setchi:2015}.

Optimal feature selection is usually unfeasible because the search space grows exponentially with the number of features. As a result various sub-optimal algorithms have been devised, with sequential forward selection being the most commonly adopted solution. Forward selection algorithms start from an empty set of features and add, in each step, the feature that jointly, i.e. together with already selected features, achieves the maximum association with the class (also called maximum relevance). Various approaches have been followed regarding how this association is accounted for.

A widely accepted association measure used in filter methods is Mutual Information (MI)~\cite{Thomas:2006}, an information-theoretic metric able to capture both linear and non-linear dependencies among random variables. One approach is to estimate directly the high-dimensional MI between the class, the already selected features, and the candidate one. However, this may not be an easy task as,  except for low dimensions, the estimation cannot rely on histograms, because of the sparse data distributions often encountered in high-dimensional spaces. 

One alternative to the estimation of high-dimensional MI or entropy measures is to use two-dimensional approximations. The usual approach is to rely on a criterion that balances the relevance of a candidate feature with its redundancy to already selected features. The relevance component is accounted through the MI between the class variable and the candidate feature. The redundancy component involves calculating the MI between the class, the already selected features, and the candidate feature. This is still a high-dimensional problem, but several approximations were considered to reduce it to two-dimensions, by assuming that the candidate and already selected features are independent, given the class ~\cite{Battiti:1994,Kwak:2002,Peng:2005,Estevez:2009,Novovicova:2007}. The methods that use this approximation are called two-dimensional since they only involve calculating the entropy between two univariate random variables.

Another class of methods has removed the conditional independence assumption of two-dimensional methods, by considering approximations involving the MI between the candidate and each selected feature, given the class. These methods are called three-dimensional since they involve calculating the entropy among three univariate variables, the class, the candidate feature and each of the already selected features. Among these, the most popular and promising methods are CIFE \cite{CIFE:2006}, JMI \cite{JMI:1999}, CMIM \cite{CMIM:2004}, and JMIM \cite{Bennasar.Setchi:2015}.

There has been an increasing concern around the evaluation of feature selection methods. The common practice is to perform the evaluation considering specific classifiers and datasets. This may explain why there are so many proposals and so little consensus on the best features to be used in particular scenarios. Filter methods are per-definition independent from the classifier and, therefore, should be evaluated independently from the classifier. This work is a first contribution in this direction. We concentrate on the analysis of two-dimensional sequential forward feature selection methods, encompassing a total of eight methods. For the analysis, we define a scenario with two classification classes and a set of representative features (relevant, redundant, and irrelevant), linearly related with the classes, which was carefully designed to bring out differences among the methods and situations where the methods may not perform correctly. A similar, but not completely coincident, scenario was considered by other authors \cite{Kwak:2002,Huang:2008}, but our analysis proceeds theoretically, to determine the true feature ordering for the methods under analysis. The ordering obtained in this way does not depend on entropy or MI estimation methods,  classifiers, or specific datasets, leading to an undoubtful comparison of the feature selection methods, which is the major advantage of our approach. Besides providing an evaluation independent from the classifier, our theoretical framework also unveils several problems intrinsic to the methods, difficult to detect through an evaluation strictly based on data. In particular, we detected inconsistencies in the construction of the objective function used to select the candidate features, due to various types of indeterminations and due to the possibility of the entropy of continuous random variables taking null and negative values.

In Section \ref{sec:entropy:MI} we review the notions of entropy and MI, as well as their properties, highlighting differences of these notions in the context of discrete and continuous distributions. Section \ref{sec:relevance_redundancy} presents the concepts of relevance and redundancy, introducing the idea of relevance-optimal sets. Section \ref{sec:feature_selection} surveys the feature selection methods that are evaluated in this work. Then, in Section \ref{sec:eval:scen}, we propose an evaluation scenario, and for that scenario derive, in Section \ref{sec:theoretical_h_mi}, the theoretical expressions of the entropies and MI that are required to obtain the true feature ordering. In Section \ref{sec:theoretical_feature_ordering}, we compare the methods under evaluation based on the true feature ordering, and discuss the shortcomings resulting from the possibility of having negative entropies and indeterminations in their objective functions. 
In Section \ref{sec:simulation} we present a simulation study on the estimation of the feature ordering which corroborates the theoretical results. Finally, in Section \ref{sec:conclusions} we draw the main conclusions of the work.

\section{Mutual information and entropy}\label{sec:entropy:MI}

MI is a measure of association between variables, capturing both linear and non-linear dependencies, that has gained wide acceptance~\cite{Thomas:2006}. The MI between two discrete random variables $X$ and $Y$, denoted $\textrm{MI}(X,Y)$, is defined by
\begin{equation}
\label{eq:mi_discrete}
\textrm{MI}(X,Y)=\sum_{x \in \mathcal{X}}\sum_{y \in \mathcal{Y}} p_{\scriptsize{XY}}(x,y) \ln{\frac{p_{\scriptsize{XY}}(x,y)}{p_X(x)p_Y(y)}}
\end{equation}
\noindent where $\mathcal{Z}$ and $p_Z(\cdot)$ denote the support and probability function of a discrete random variable or random vector $Z$, with the convention that $0\ln 0=0$. It follows from the definition that
\vspace*{1mm}
\begin{enumerate}
	\item[(a)] $\textrm{MI}\left(X,Y\right)=\textrm{MI}\left(Y,X\right)$.
	\item[(b)] $\textrm{MI}\left(X,Y\right)\geq0$, with equality for independent random variables.
\end{enumerate}

The MI between $X$ and $Y$ can also be written in terms of the entropies of $X$ and $Y$. The entropy of a discrete random variable $Z$, $H(Z)$, is a measure of the uncertainty of $Z$ and is given by
\begin{equation}
\label{eq:entropy_discrete}
H(Z)=-\sum_{z \in \mathcal{Z}} p_Z(z)\ln p_Z(z).
\end{equation}
\noindent The definition of entropy can be extended to two or more discrete random variables. For the case of two discrete random variables $X$ and $Y$, the entropy of $(X,Y)$ is defined by:
\begin{equation}
H(X,Y)=-\sum_{x \in \mathcal{X}}\sum_{y \in \mathcal{Y}} p_{\scriptsize{XY}}(x,y)\ln p_{\scriptsize{XY}}(x,y).
\end{equation}
It follows that entropy is a nonnegative function, which is null only for degenerate (point mass) random variables or vectors. After performing simple analytical manipulations, one may conclude that
\begin{enumerate}\label{eq:mi_h_discreta}
	\item[(c)] $\textrm{MI}\left(X,X\right)=H(X)$;
	\item[(d)] $\textrm{MI}\left(X,Y\right)=H(X)+H(Y)-H(X,Y)$;
	\item[(e)] $\textrm{MI}\left(X,Y\right)=H(X)-H(X|Y)$;
\end{enumerate}
\noindent where the conditional entropy of $X$ given $Y$, $H(X|Y)$, is given by
\begin{equation}
\label{eq:entropy_cond_discreta}
H(X|Y)=-\sum_{x \in \mathcal{X}} \sum_{y \in \mathcal{Y}} p_{\scriptsize{XY}}(x,y) \ln p_{X|Y=y}(x).
\end{equation}

The MI between continuous random variables $X$ and $Y$ is defined similarly to the case of discrete random variables. In detail, if we let $f_{Z}(\cdot)$ denote the probability density function of a random variable or random vector $Z$, then for $(X,Y)$ (absolutely) continuous:
\begin{equation}
\label{eq:mi_continuous}
\textrm{MI}(X,Y)=\int_\mathcal{Y} \int_\mathcal{X} f_{\scriptsize{XY}}(x,y) \ln{\frac{f_{\scriptsize{XY}}(x,y)}{f_{X}(x) f_{Y}(y)}} \ud x \, \ud y.
\end{equation}
\noindent Note that properties (a)-(b) stated above also hold for (absolutely) continuous random pairs $(X,Y)$. Likewise the MI, the entropy of $X$, $h(X)$, the entropy of $(X,Y)$, $h(X,Y)$, and the conditional entropy of $X$ given $Y$, $h(X|Y)$, are given by:
\begin{align}
\label{eq:entropy_continuous}
h(X)&= \,-\int_{\mathcal{X}} f_X(x)\ln f_X(x) \ud x,\\
h(X,Y)&=\,-\int_{\mathcal{Y}} \int_{\mathcal{X}} f_{\scriptsize{XY}}(x,y)\ln f_{\scriptsize{XY}}(x,y)\, \ud x\, \ud y,\\
h(X|Y)&=\,-\int_{\mathcal{Y}} \int_{\mathcal{X}} f_{\scriptsize{XY}}(x,y) \ln f_{X|Y=y}\left(x\right) \, \ud x\, \ud y.
\end{align}
In the remainder of this section, we will use the common terminology of calling differential entropy the entropy function for continuous random variables, $h(\cdot)$. In the paper, we will drop the term ``differential'' whenever it is clear that we are referring to continuous random variables.

It is important to note that entropy and differential entropy do not share the same properties, even though properties (d)-(e) above hold with entropy substituted by differential entropy.
For example, contrarily to the entropy, which is always nonnegative, the differential entropy can take both positive and negative values, as well as zero. This fact is nicely illustrated, for example, by the uniform distribution on the interval $[0,a]$, $a>0$, $\textrm{Unif}(0,a)$, for which
$$
X \sim \textrm{Unif}(0,a) \Longrightarrow h(X)=\ln a.
$$
Thus, $h(X)$ is positive (null, negative) if $a>1$ ($a=1$, $a<1$). Note, in particular, that the uniform distribution on the interval $[0,1]$ has null differential entropy despite this distribution not being close to a degenerate one. Another property that is not shared by entropy and differential entropy is $\textrm{MI}\left(X,X\right)$ being equal to the entropy of $X$, which we have seen to hold for a discrete random variables [see property (c) above]. A very different result is obtained in the case of an absolutely continuous random variable $X$, namely that $\textrm{MI}\left(X,X\right)=+\infty$, as stated in \cite{Kotz:1966} and \cite{Brillinger:2004}.

In practice, it is also important to compute the MI  between discrete and continuous random variables. This is the case, for example, in feature selection problems involving a continuous candidate feature and a discrete class variable. For continuous $X$ and discrete $Y$ random variables, the MI between $X$ and $Y$ is given by
\begin{equation}\label{eq:mi_cont_disc}
\textrm{MI}\left(X,Y\right)=\sum_{y \in \mathcal{Y}} \int_{\mathcal{X}} f_{X|Y=y}(x) p_Y(y) \ln \frac{f_{X|Y=y}(x)}{f_X(x)} \, \ud x.
\end{equation}
\noindent One may note that properties (a)-(b) and the analogous of properties (d)-(e) still hold in this case.

When dealing with more than two variables, it arises the need to compute the MI among three or more variables. One of the main definitions of MI among three variables is the triple mutual information, TMI~\cite{Triple:1995}. For example, the TMI among  continuous random variables $X$, $Y$, $Z$, with joint probability density function $f(x,y,z)$, and marginal distributions $f_{XY}(x,y)$, $f_{XZ}(x,z)$, $f_{YZ}(y,z)$, $f_{X}(x)$, $f_{Y}(y)$, and $f_{Z}(z)$ is given by
%
%
\begin{flalign*}
{\rm TMI}(X,Y,Z)=\int_{\mathcal{Z}} \int_{\mathcal{Y}} \int_{\mathcal{X}} f(x,y,z) \ln\frac{f_{XY}(x,y)f_{XZ}(x,z)f_{YZ}(y,z)}{f(x,y,z)f_X(x)f_{Y}(y)f_{Z}(z)}
\ud x \ud y \ud z.
\end{flalign*}
%
\noindent
Using this definition, we can prove that for random variables $X$ and $Y$ and a random variable or random vector $\boma{Z}$:
\begin{eqnarray}\label{MI_Chain_Rule}
\textrm{TMI}\left(X,Y,\boma{Z}\right)&=&\textrm{MI}\left(X,Y\right)-\textrm{MI}\left(X,Y|\boma{Z}\right)\nonumber\\
&=&\textrm{MI}\left(X,\boma{Z}\right)-\textrm{MI}\left(X,\boma{Z}|Y\right)\\
&=&\textrm{MI}\left(Y,\boma{Z}\right)-\textrm{MI}\left(Y,\boma{Z}|X\right),\nonumber
\end{eqnarray}
where
\begin{displaymath}
\textrm{MI}\left(X,Y|\boma{Z}\right)= h(X|\boma{Z})-h(X|Y,\boma{Z}),
\end{displaymath}
and analogously for the other cases. The definition of $\textrm{TMI}(X,Y,\boma{Z})$ has the disadvantage of assuming not only positive or null values but also negative ones~\cite{Triple:1995}, which demands a new interpretations of MI. 

\section{Relevance and redundancy}
\label{sec:relevance_redundancy}

Feature selection methods share the general goal of identifying an appropriate subset of the original features with the property of being {\em maximally informative} about the class~\textcolor{blue}{\cite{Vergara.Estevez:2014,Huang:2008}}. Following the principle of parsimony, among the maximally informative sets the ones that have minimum size are to be preferred; we call these minimum size sets {\em relevance-optimal} sets.

In order to introduce the notion of maximally informative and relevance-optimal sets, it is convenient to introduce some notation. We let $\boma{V} =(V_i)_{i\in T}$ denote the set of all input features and $C$ the class (random variable). Moreover, for a subset $L$ of $T$ we let $\boma{V}_L =(V_l)_{l\in L}$, and similarly for an observation $\boma{v}=(v_i)_{i\in T}$ of $\boma{V}=(V_i)_{i\in T}$ we let $\boma{v}_L=(v_l)_{l\in L}$. In addition, we let $\overline{L}=T\setminus L$ denote the complement of $L$, $T_{-i}=T\setminus \{i\}$, and $\stackrel{\mathrm{d}}{=}$ denote equality in distribution.
\begin{definition}
	\label{def:max_relevant}
	The feature set $\boma{V}_L$ is maximally informative (for class $C$) if for all $\boma{v}$ in the support of $\boma{V}$,
	$$
	\left[C|(\boma{V}_L,\boma{V}_{\overline{L}})=(\boma{v}_L,\boma{v}_{\overline{L}})\right] \stackrel{\mathrm{d}}{=}
	\left[C|\boma{V}_L=\boma{v}_L\right].
	$$
	Moreover, a maximally informative (feature) set is a relevance-optimal set if it has minimum size among all maximally informative sets.
\end{definition}
Thus, a feature set $\boma{V}_L$ is maximally informative for class $C$ if knowledge on features not belonging to $\boma{V}_L$ does not impact the conditional distribution of $C$, provided the values of the features belonging to $\boma{V}_L$ are known. With the previous definition, we are in condition to introduce the concept of {\em irrelevant} feature, as well as two concepts of feature relevance: {\em strongly relevant} feature and {\em weakly relevant} feature. 

\begin{definition}
	\label{def:SR}
	A feature $V_i$ is strongly relevant if $\boma{V}_{T_{-i}}$ is not maximally informative and is irrelevant if for all
	$L\subseteq T_{-i}$ and all $(v_i,\boma{v}_L)$ in the support of $(V_i,\boma{V}_L)$:
	$$
	\left[C|(V_i,\boma{V}_L)=(v_i,\boma{v}_L)\right] \stackrel{\mathrm{d}}{=}
	\left[C|\boma{V}_L=\boma{v}_L\right].
	$$
	A feature that is neither strongly relevant nor irrelevant is called weakly relevant.
\end{definition}
The previous definition leads to a partition of the set of features into strongly relevant (SR), weakly relevant (WR), and irrelevant features, with the definitions of SR, WR, and irrelevant features coinciding with the ones presented in \cite{John.Kohavi.Pfleger:1994} and \cite{Yu.Liu:2004}. For a characterization of SR feature, WR feature, and irrelevant feature based on TMI see \cite{Brown:2012}.

Note that a SR feature belongs to all relevance-optimal sets. Conversely, an irrelevant feature belongs to no relevance-optimal set. Furthermore, a relevance-optimal set may either contain or do not contain a specific WR feature. Thus, in general the identification of relevant (SR and WR) features is not enough to get a relevance-optimal subset, since duplications or other kinds of functional dependencies may occur among WR features. The next example, which is inspired in Example 1 of \cite{Yu.Liu:2004}, illustrates this fact.

\begin{exemplo}
	\label{ex:optimal_subset}
	Let $T=\{1,2,\ldots,5\}$, with features $V_1$, $V_2$, and $V_4$ being independent, $V_3=3\,V_2+1$, and $V_5=(V_4)^2$. Moreover, assume that $C=g(V_1,V_2)$ is a binary random variable such that $g(v_1,v_2)$ is not constant in either of the variables $v_1$ and $v_2$.
	
	The sets containing feature $V_1$ and one of the features $V_2$ and $V_3$ are maximally informative since both $(V_1,V_2)$ and $(V_1,V_3)$ determine $C$, whereas none of the features $V_1$, $V_2$, and $V_3$ in isolation determines $C$. As a by-product, we conclude that: $V_1$ is the unique SR feature; $V_2$ and $V_3$ are WR features; and $V_4$ and $V_5$ are irrelevant features. Moreover, there are two relevance-optimal sets: $(V_1,V_2)$ and $(V_1,V_3)$.
	
	Note that the relevance-optimal sets have size two, thus implying that a  minimum of two features are needed to convey all information on the class that is contained in $(V_1,V_2,\ldots,V_5)$.
\end{exemplo}

\noindent The scientific community quickly realized that, for the large and complex feature sets commonly found in practice, it may be impractical to derive all relevance-optimal sets. This has paved the way to the development of the systematic approach to derive (just) a single relevance-optimal set using Markov blanket filtering~\cite{Koller.Sahami:1996}.

Markov blanket filtering is a backward elimination process that starting from the set of relevant (SR and WR) features, say $\boma{V}_R$, eliminates one by one WR features until a relevance-optimal set is obtained. Each step of the backward elimination process consists in selecting a feature $V_j$ from the current maximally-relevant subset $\boma{V}_N$ of $\boma{V}_R$ for which there exists a Markov blanket $\boma{V}_M$, $M\subseteq N\setminus\{j\}$, meaning that for any $(v_j,\boma{v}_M)$ in the support of $(V_j,\boma{V}_M)$,
$$
\left[(C,\boma{V}_K)|(V_j,\boma{V}_M)=(v_j,\boma{v}_M)\right] \stackrel{\mathrm{d}}{=}
\left[(C,\boma{V}_K)|\boma{V}_M=\boma{v}_M\right].
$$
with $K=N\setminus (M\cup \{j\})$. Following this way, a relevance-optimal set is obtained when none of the features $V_j$, $j\in N$, possesses a Markov blanket.

Yu and Liu in \cite{Yu.Liu:2004} rightly pointed out that we cannot find a  Markov blanket for strongly relevant features, thus implying that a relevance-optimal set contains necessarily all SR features. However, a relevance-optimal set contains only a part of the WR features, as illustrated in the example above. As a result, each relevance-optimal set leads naturally to the following classification of WR features in two types: WR features that belong to the relevance-optimal set, and WR features that do not belong to the relevance-optimal set (see~\cite{John.Kohavi.Pfleger:1994,Yu.Liu:2004}, and references therein). Following \cite{Yu.Liu:2004}, we call the former weakly relevant and non-redundant (WR-NR) features and the latter weakly relevant and redundant (WR-R) features. As a result, one gets a partition of the set of features in four subsets: SR features, WR-NR features, WR-R features, and irrelevant features.

One should stress that the partition of the features in four sets (SR, WR-NR, WR-R, and irrelevant) thus obtained is a function of the relevance-optimal set used to divide WR features into WR-NR and WR-R features. As an illustration, note that in Example~\ref{ex:optimal_subset} the feature $V_2$ is WR-NR for the relevance-optimal set $(V_1,V_2)$ and WR-R for the relevance-optimal set $(V_1,V_3)$, with the converse holding for feature $V_3$.

To end the section, we remark that not all relevance-optimal sets should be considered equally good from a practical point of view. In fact, the degrees of redundancy (or association) between the features of different relevance-optimal sets are not necessarily equal, in which case relevance-optimal sets whose features exhibit the lowest level of redundancy should be preferred. This may be interpreted as a reason for selecting a relevance-optimal minimum-redundancy feature set.

\section{Feature selection methods}\label{sec:feature_selection}

In feature selections problems, the practitioner aims at finding features that contain as much information as possible about the class variable while, following the principle of parsimony and seeking to improve interpretability, avoid selecting features that contain redundant information with respect to the class variable. We will concentrate in the framework of forward sequential methods that progressively select a new feature, to add to the set of selected features, using criteria based on MI measures.

With $S$ ($F$) denoting the set of already selected (unselected) features at a given step, $F\cap S=\emptyset$, the concrete objective turns out to be the selection of an additional feature $V_i\in F$ such that the MI between the class variable ($C$) along with the already selected features ($S$) and the candidate feature ($V_i$), $\textrm{MI}(C,\{S,V_i\})$, is maximized. However, 
\begin{align*}
\textrm{MI}(C,\{S,V_i\}) &= \textrm{MI}(C,S) + \textrm{MI}(C,V_i) - \textrm{TMI}(C,V_i,S)\\
&=\textrm{MI}(C,S) + \textrm{MI}(C,V_i|S),
\end{align*}
and $\textrm{MI}(C,S)$ does not depend on $i$, this amounts to pick a feature $V_i\in F$ such that the conditional MI between $V_i$ and the class variable given the set of already selected features, $\textrm{MI}(C,V_i|S)$, is maximized. 
Here $\textrm{TMI}(C,V_i,S)$ represents the redundancy between the candidate feature $V_i$, the already selected features $S$, and the class variable $C$. 
As a result, the decomposition of $\textrm{MI}(C,V_i|S)$ 
expresses a trade off  between the {\em relevance} of the candidate feature to explain the class variable, encompassed in $\textrm{MI}(C,V_i)$, and its {\em redundancy} for the same effect in face of the previously selected features, $\textrm{TMI}(C,V_i,S)$.

One should note that in practice the estimation of $\textrm{TMI}(C,V_i,S)$ is problematic even in the case where $S$ contains a small number of features~\cite{Kwak:2002,Peng:2005,Huang:2008}. Accordingly, several feature selection methods that use simplifications to approximate $\textrm{TMI}(C,V_i,S)$ have been introduced in the literature. In this paper, we only study methods that consider a two-dimensional approximation of the TMI (for three-dimensional alternatives see \cite{Brown:2012,Vergara.Estevez:2014,Bennasar.Setchi:2015, Freeman.et.al:2015, Vinh.et.al:2016} and references therein).
These feature selection methods include: MIFS~\cite{Battiti:1994}, MIFS-U~\cite{Kwak:2002}, mRMR~\cite{Peng:2005}, mMIFS-U~\cite{Novovicova:2007}, MICC~\cite{Huang:2008}, QMIFS~\cite{Huang:2008}, and NMIFS~\cite{Estevez:2009}. In brief, these methods select at each step a feature according to the following type of criteria
\begin{equation}
\label{gencriteria}
\argmax_{V_i \in F} \left\{\textrm{MI}\left(C,V_i\right)-\overline{\textrm{TMI}}\left(C,V_i,S\right)  \right\}
\end{equation}
where $\overline{\textrm{TMI}}\left(C,V_i,S\right)$ is a method dependent approximation to $\textrm{TMI}(C,V_i,S)$. We next present the specific forms of $\overline{\textrm{TMI}}\left(C,V_i,S\right)$  for the mentioned methods. The objective functions for the methods are summarized in Table \ref{tab:feature_selection_methods}.

\begin{table*}[!h]
	\centering
	\scriptsize
	\caption{Objective functions of the eight different feature selection methods: MIFS, MIFS-U, mRMR, mMIFS-U, QMIFS, MICC, NMIFS  and maxMIFS.}
	\label{tab:feature_selection_methods}
	\begin{tabular}{lc}
		\toprule
		Method & Objective function \\
		\midrule\\
		MIFS~\cite{Battiti:1994} & $ \dps \argmax_{V_i \in F}\left\{\textrm{MI}\left(C,V_i\right)-\beta \sum_{V_s \in S} \textrm{MI}\left(V_i,V_s\right)  \right\} $ \\[1cm]
		MIFS-U~\cite{Kwak:2002}  &  $ \dps \argmax_{V_i \in F}\left\{\textrm{MI}\left(C,V_i\right)-\beta \sum_{V_s \in S} \frac{\textrm{MI} \left(C,V_s\right)}{h(V_s)} \textrm{MI}\left(V_i,V_s\right) \right\}$ \\[1cm]
		mRMR~\cite{Peng:2005}    & $ \dps \argmax_{V_i \in F}\left\{ \textrm{MI}\left(C,V_i\right) -\frac{1}{\left|S\right|} \sum_{V_s \in S} \textrm{MI} \left(V_i,V_s\right)\right\} $\\[1cm]
		mMIFS-U~\cite{Novovicova:2007} &  $ \dps \argmax_{V_i \in F}\left\{ \textrm{MI}\left(C,V_i\right) - \max_{V_s \in S} \frac{\textrm{MI}\left(C,V_s\right)}{h(V_s)} \textrm{MI}\left(V_i,V_s\right)\right\}$ \\[1cm]
		MICC~\cite{Huang:2008} & $ \dps \argmax_{V_i \in F}\left\{\frac{\textrm{MI}\left(C,V_i\right)}{ \dps \frac{1}{\left|S\right|} \sum_{V_s \in S} \frac{\textrm{MI}\left(V_i,V_s\right)}{\min\left\{h(V_i),h(V_s)\right\}}}-
		\textrm{MI}\left(C,V_i\right)  \right\} $\\[1cm]
		QMIFS\cite{Huang:2008} & \hspace*{-0.8cm} $ \dps \argmax_{V_i \in F}\left\{\textrm{MI}\left(C,V_i\right)-\sum_{V_k \in S}
		\left(\frac{\textrm{MI}\left(V_i,V_k\right)}{h(V_k)}-\frac{1}{2}\sum_{\substack{V_j \in S \\ j\neq k}} \frac{\textrm{MI}\left(V_i,V_j\right)}{h(V_j)} \frac{\textrm{MI}\left(V_j,V_k\right)}{h(V_k)} \textrm{MI}\left(C,V_k\right)\right)\right\}$ \\[1cm]
		NMIFS~\cite{Estevez:2009}& $ \dps \argmax_{V_i \in F}\left\{\textrm{MI}\left(C,V_i\right)-\frac{1}{\left|S\right|} \sum_{V_s \in S} \frac{\textrm{MI}\left(V_i,V_s\right)}
		{\min\left\{h(V_i),h(V_s)\right\}}\right\}$\\[1cm]
		maxMIFS~\cite{Pascoal:2014} & $ \dps \argmax_{V_i \in F}\left\{\textrm{MI}\left(C,V_i\right)-\max_{V_s \in S}\left\{\textrm{MI}\left(V_i,V_s\right)\right\}  \right\}$\\[0.7cm]
		\bottomrule
	\end{tabular}
\end{table*}
The first proposal, developed in \cite{Battiti:1994} and called MIFS method, uses the approximation
\begin{equation}
\label{Battitiapprox}
\overline{\textrm{TMI}}\left(C,V_i,S\right) = \beta \sum_{V_s\in S} \textrm{MI}(V_i,V_s)
\end{equation}
where $\beta\in[0,1]$ is a weight factor that should be chosen by the user. One may arrive at this approximation by introducing a weight factor $\beta$ after initially assuming that:
\begin{itemize}
	\item[(a)] $\textrm{TMI}\left(C,V_i,S\right)=\textrm{MI}\left(V_i,S\right)$.
	\item[(b)] The already selected features are independent.
\end{itemize}
Battiti's \cite{Battiti:1994} first assumption states that, given a certain class $C=c$, the candidate feature, $V_i$, and the already selected features, $S$, are independent, an hypothesis of conditional independence. The assumptions (a)-(b) lead to
$$
\textrm{TMI}\left(C,V_i,S\right) = \textrm{MI}\left(V_i,S\right) = \sum_{V_s\in S} \textrm{MI}(V_i,V_s).
$$

The introduction of the weight factor $\beta$ may thus be regarded as a correction factor for deviations from the two mentioned assumptions. This parameter was viewed by the author of MIFS as regulating the relative importance of the redundancy component. \cite{Battiti:1994} claimed that a value for $\beta$ [in (\ref{Battitiapprox})] between 0.5 and 1 is appropriate for many classification tasks. However, several authors have argued that the best choice for $\beta$ being problem dependent constitutes an important drawback of MIFS.

The mRMR method, proposed in \cite{Peng:2005}, avoids  the need to choose a value for the parameter $\beta$. Even though it has been derived by its authors as a criteria combining maximum relevance with minimum redundancy, the mRMR method corresponds to a variation of the MIFS method through the introduction of an adaptive $\beta$ that evolves as the number of already selected features changes, being effectively the reciprocal of the the number of already selected features. More precisely, mRMR uses the approximation
\begin{equation}
\label{redmRMR}
\overline{\textrm{TMI}}\left(C,V_i,S\right) = \frac{1}{|S|} \sum_{V_s\in S}
\textrm{MI}(V_i,V_s).
\end{equation}
Note that the redundancy component associated with the selection of candidate feature $V_i$ is here measured by the mean of the MI between $V_i$ and each of the already selected features, $V_s \in S$.

With the aim of addressing the fact that the entropy of random variables may vary greatly, \cite{Estevez:2009} claims that the MI values between the candidate feature and the already selected features should be normalized. Accordingly, its authors proposed to substitute $\textrm{MI}(V_i,V_s)$ by $\textrm{NI}(V_i,V_s)$, the normalized mutual information between the features $V_i$ and $V_s$, given by
\begin{equation}
\label{normMI}
\textrm{NI}(V_i,V_s) = \frac{\textrm{MI}(V_i,V_s)}{\min{\left\{h(V_i),h(V_s)\right\}}}.
\end{equation}
In sequence, they proposed in the same paper the NMIFS method, which uses the approximation
\begin{equation}
\label{redmRMR}
\overline{\textrm{TMI}}\left(C,V_i,S\right) = \frac{1}{|S|} \sum_{V_s\in S}
\textrm{NI}(V_i,V_s).
\end{equation}
The use of $\textrm{NI}(V_i,V_s)$ in NMIFS -- instead of $\textrm{MI}(V_i,V_s)$, like in mRMR -- as a measure of redundancy between the candidate feature $V_i$ and the already selected feature $V_s$ was justified with the supposed fact that
$$
0\leq \textrm{MI}(V_i,V_s)\leq \min\left\{h(V_i),h(V_s)\right\}
$$	
leading to $0\leq \textrm{NI}(V_i,V_s)\leq 1$. However, the second inequality in the above equation only holds with certainty when $V_i$ and $V_s$ are discrete random variables.
In fact, as the entropies of continuous random variables may take negative values, $\textrm{NI}(V_i,V_s)$ can take negative values, leading to the redundancy of $V_i$ with respect to $V_s$ being weighted positively, contrarily to what was intended. This problem extends to all other methods that incorporate entropies of features in denominators of fractions.

Kwak and Choi~\cite{Kwak:2002} introduced the MIFS-U method, whose basis is similar to that of MIFS, but where the authors tried to overcome the assumption of independence between the class and the redundancy component - assumption (a), while maintaining the independence assumption for the already selected features - assumption (b). Specifically, MIFS-U uses the approximation
\begin{equation}
\label{redMIFSU}
\overline{\textrm{TMI}}\left(C,V_i,S\right) = \beta \sum_{V_s\in S}
\frac{\textrm{MI}(C,V_s)}{h(V_s)}\, \textrm{MI}(V_i,V_s)
\end{equation}
by assuming that the class variable does not change the ratio of the MI of the candidate feature with a single already selected feature to the entropy of that already selected feature, i.e.,
\begin{equation}
\label{proprMIE}
\frac{\textrm{MI}(V_i,V_s)}{h(V_s)} = \frac{\textrm{MI}(V_i,V_s|C)}{h(V_s|C)}
\end{equation}
for each $V_s\in S$. A direct consequence of this assumption is that
\begin{align*}
\textrm{TMI}(C,V_i,V_s) &= \textrm{MI}(V_i,V_s) - \textrm{MI}(V_i,V_s|C)\\
&= \left[1- \frac{h(V_s|C)}{h(V_s)}\right] \textrm{MI}(V_i,V_s)\\
&= \frac{\textrm{MI}(C,V_s)}{h(V_s)}\, \textrm{MI}(V_i,V_s),
\end{align*}
which leads to the term that appears in (\ref{redMIFSU}), the MIFS-U approximation for
$\textrm{MI}\left(C,V_i,S\right)$.

The assumption (\ref{proprMIE}) is somehow counterintuitive as one expects that if features are associated with the class variable, then knowledge of the class variable would lead to different conditional information on the features. Moreover the appearance of the entropies of the already selected features in the denominators of fractions in (\ref{redMIFSU}) constitutes a drawback of MIFS-U in the presence of already selected features with entropy close to zero, and especially in the presence of already selected continuous-type features with negative entropy. As a result, the approximation (\ref{redMIFSU}) for $\textrm{TMI}\left(C,V_i,S\right)$ may turn out to be negative, leading to the redundancy of the candidate feature with already selected features being weighted positively, contrarily to what was desired.

Novovicov{\'a} and co-authors~\cite{Novovicova:2007} proposed the mMIFS-U method, which uses the approximation
\begin{equation}
\label{redmMIFSU}
\overline{\textrm{TMI}}\left(C,V_i,S\right) =  \max_{V_s\in S}
\left\{\frac{\textrm{MI}(C,V_s)}{h(V_s)}\, \textrm{MI}(V_i,V_s)\right\}.
\end{equation}
Like the MIFS-U method, mMIFS-U assumes the condition (\ref{proprMIE}), and shares with MIFS-U the drawbacks resulting from having entropies of already selected variables appearing in the denominators of fractions. Conversely, contrarily to MIFS-U, mMIFS-U avoids the problem of selecting an appropriate value for $\beta$  by replacing a sum over the already selected features in (\ref{redMIFSU}) by a maximum over the same set of features in (\ref{redmMIFSU}).

Later, \cite{Huang:2008} introduced the QMIFS method, which uses the following approximation for $\textrm{MI}\left(C,V_i,S\right)$ with the aim of incorporating possible interactions between two (but not more than two) already selected features:
\begin{equation}
\label{redQMIFS}
\overline{\textrm{MI}}\left(C,V_i,S\right) =
\sum_{V_k \in S}
\left[\phi_{ik}-\frac{1}{2}  \sum_{\substack{V_j \in S \\ j\neq k}}
\phi_{ij}\phi_{jk}\right]\textrm{MI}(C,V_k)
\end{equation}
where $\dps \phi_{lm}=\textrm{MI}(V_l,V_m)/h(V_m)$ for $l\neq m$.
The presentation of the authors for the derivation of this approximation is not easy to follow and seems to be based on several assumptions that may be hardly satisfied in practice. In particular, aside from the condition (\ref{proprMIE}), the authors assume the following property on the information of the candidate feature $V_i$ and pairs of already selected features $(V_j,V_k)$, with $V_j\neq V_k$:
\begin{equation}
\label{proprQMIFS}
\frac{\textrm{MI}(V_i,V_j)}{h(V_j)} = \frac{\textrm{MI}(V_i,V_j,V_k)}{\textrm{MI}(V_j,V_k)}.
\end{equation}
As a result of (\ref{redQMIFS}) and what has been stated, one concludes that QMIFS shares the main drawbacks of MIFS-U not related with the parameter $\beta$ of the latter.

In this paper we consider an additional feature selection method, which we call maxMIFS. This method is similar to mRMR~\cite{Peng:2005}, but uses the maximum MI between the candidate feature and individual already selected features instead of their mean. That is, maxMIFS is a method of the generic type criteria (\ref{gencriteria}) with the approximation
\begin{equation}
\label{redROPAmax}
\overline{\textrm{TMI}}\left(C,V_i,S\right) = \max_{V_s\in S} \textrm{MI}(V_i,V_s).
\end{equation}
Note that the use of the maximum of the MI between the candidate feature and each of the already selected features avoids overweighting the redundancy component of the objective function.

To end the section, one should mention that \cite{Huang:2008} proposed a second feature selection method called MICC. Like NMIFS, this method is based on the use of the normalized mutual information between the candidate features $V_i$ and already selected $V_s$, $\textrm{NI}(V_i,V_s)$, instead of the MI between the same variables. In detail, MICC uses as criteria for selecting a new feature the candidate feature $V_i$ that maximizes the following expression:
\begin{equation}
\label{redMICC}
\overline{\textrm{TMI}}\left(C,V_i|S\right) = \textrm{MI}\left(C,V_i\right)
\left[\frac{|S|}{\sum_{V_s\in S} \textrm{NI}(V_i,V_s)} - 1 \right].
\end{equation}
Similarly to the reasoning followed in the proposal of the NMIFS method~\cite{Estevez:2009}, it is claimed in \cite{Huang:2008} that the multiplicative factor affecting $\textrm{MI}\left(C,V_i\right)$ in the previous equation takes values on $[0,\infty]$. However, this conclusion may be false for continuous features.

\section{Evaluation scenario}\label{sec:eval:scen}

\noindent Evaluating feature selection methods can be done in two ways. The first one is to embed the classifier in the evaluation process~\cite{Battiti:1994,Peng:2005,Novovicova:2007}. In this case, the methods are compared based on the accuracy of the classification process estimated using labeled data (data for which the true class in known). The results obtained with this method are difficult to generalize, since they depend on the specific classifier and on the performance metrics used in the comparison. The second evaluation is based on scenarios defined by an initial set of \textit{interesting} features and a relation between these features and the output class~\cite{Kwak:2002,Huang:2008}. In this case, the true ordering of features must be known, and the methods are compared based on how well they can approach it. A reference that may be used in this type of evaluation is the one obtained with the Markov blanket filtering methodology described in Section \ref{sec:relevance_redundancy}. In this work we will concentrate on the last type of evaluation.

There are three requirements that a good evaluation scenario must observe. First, it must be challenging, i.e., it must lead to situations where the decision metrics used in selecting candidate features are close enough to favor wrong decisions. Second, it must include a representative set of features, containing relevant, redundant, and irrelevant ones. Finally, it must be amenable to theoretical evaluation, i.e., one should be able to obtain the \textit{true} ordering of features for the methods under analysis. If this last requirement is not fulfilled, the evaluation can only be based on a conjecture of what the true ordering is, which may lead to erroneous evaluation. To the best of our knowledge, our work is the first one to utilize a theoretical framework in the evaluation of feature selection methods.

Kwak and Choi~\cite{Kwak:2002} and Huang et al.\cite{Huang:2008} proposed an evaluation scenario with two classes defined by
\begin{equation}\label{eq:C:QMIFS}
C = \left\{ \begin{array}{rl}
0, &\mbox{$X+0.2Y<0$} \\
1, &\mbox{$X+0.2Y\geq0$}
\end{array} \right.
\end{equation}

\noindent where $X$ and $Y$ are independent random variables uniformly distributed in $\left[-0.5,0.5\right]$. \cite{Kwak:2002} considered as interesting features $X$, $Y^2$, and $X-Y$;  \cite{Huang:2008} added seven other features, including $Z$ and functions of it, where $Z$ is independent and identically distributed to $X$ and $Y$ (see Table \ref{tab:input:features:Huang} for the complete list). Both evaluation scenarios are amenable to theoretical evaluation, as it will become clear in next section, but the authors did not pursue this goal.
\begin{table}[!h]
	\centering
	\scriptsize
	\caption{Input features proposed in \cite{Huang:2008}.}
	\begin{tabular}{ccccc}
		\toprule
		Features & Description  & & Features & Description \\
		\midrule
		$V_1$ & $X$    & & $V_6$ & $Z^2$\\[0.1cm]
		$V_2$ & $3X+1$ & & $V_7$ & $ZX$\\[0.1cm]
		$V_3$ & $Y^2$ & & $V_8$ & $ZY$\\[0.1cm]
		$V_4$ & $X-Y$  & & $V_9$ & $ZXY$\\[0.1cm]
		$V_5$ & $Z$  & & $V_{10}$ & $ZY$\\
		\bottomrule
	\end{tabular}
	\label{tab:input:features:Huang}
\end{table}

Using the framework of Section \ref{sec:relevance_redundancy}, the features in \cite{Huang:2008} can be classified in the following way: there are no strongly relevant features; $Z$ and $Z^2$ are irrelevant; $Y^2$ and $ZXY$ are always WR-R features. Moreover, there are two relevan\-ce-optimal sets: $(X,X-Y)$ and $(3X+1,X-Y)$.  
This selection of features deserves the following comments:

\begin{itemize}
	\item It is reasonable that no strongly relevant feature has been included, since these have a high probability of being selected as relevant. SR features do not put the feature selection method under stress.
	\item The number of interesting relevance-optimal sets is too small. In fact, as discussed in Section \ref{sec:feature_selection}, the methods under analysis perform selection by evaluating the relevance of features (to the class) and the redundancy between the candidate and already selected features. Thus, it is important to include in the initial set features that lead to relevance-optimal sets with different levels of redundancy among features.
	\item Strangely, $Y$ was not included in the set of features, given that it is one of the features used in the class definition. Including $Y$ would have added two relevance-optimal sets, $(X,Y)$ and $(3X+1,Y)$, where features are independent among themselves. Moreover, to evaluate how well the feature selection methods match the true feature ordering, it is important to confront the possibility of selecting $(X,Y)$ or $(X,X-Y)$, or equivalently $(3X+1,Y)$ and $(3X+1,X-Y)$. These two outcomes are easily confused. Indeed, as we will show latter, $X-Y$ has a MI with the class which is larger than that of $Y$. Thus, depending on the relative strength of the redundancy component, either $(X,Y)$ or $(X,X-Y)$ may be selected first.
\end{itemize}

Based on the above comments, we generalized the evaluation scenario of \cite{Huang:2008}, in the following way. First, we included $Y$ in the set of features. Second, we removed features $ZX$, $ZY$, and $ZXY$, because theoretical analysis is involved and these are necessarily WR-R features. Finally, we added two irrelevant features $W$ and $W+Z$, to assess whether the feature selection methods lead to particular patterns of feature ordering (e.g. irrelevant or redundant features following the relevance-optimal set). Our scenario is then based on the $10$ features shown in Table \ref{tab:input_features}. We also expanded the class definition, to contemplate different relative strengths between $X$ and $Y$. Specifically, the two classes are defined by
\begin{eqnarray}\label{eq:class_Ck}
C_k = \left\{ \begin{array}{rl}
0, &\mbox{$X+kY<0$} \nonumber\\
1, &\mbox{$X+kY\geq0$}
\end{array} \right.,
\end{eqnarray}
\noindent where $k \in \left[0,1\right]$. In this way, our scenario has four irrelevant features, $Z$, $Z^2$, $W+d$, and $W+Z$, no strongly relevant feature, two features that are WR-R, $X^2$ and $Y^2$, and five relevance-optimal sets, $(X,Y)$, $(X,X-Y)$, $(Y,X-Y)$, $(3X+1,Y)$, and $(3X+1,X-Y)$.
\begin{table}[!h]
	\centering
	\scriptsize
	\caption{Input features of evaluation scenario.}
	\begin{tabular}{ccccc}
		\toprule
		Features & Description  & & Features & Description \\
		\midrule
		$V_1$ & $X$    & & $V_6$    & $Z^2$\\[0.1cm]
		$V_2$ & $aX+b$ & & $V_7$    & $Y$\\[0.1cm]
		$V_3$ & $Y^2$  & & $V_8$    & $X^2$\\[0.1cm]
		$V_4$ & $X-Y$  & & $V_9$    & $W+d$\\[0.1cm]
		$V_5$ & $Z$    & & $V_{10}$ & $Z+W$\\[0.1cm]
		\bottomrule
	\end{tabular}
	\label{tab:input_features}
\end{table}

\section{Theoretical entropy and mutual information}\label{sec:theoretical_h_mi}

In this section we summarize the theoretical results needed to compare the feature selection methods. We consider two different scenarios, where the random variables $X$, $Y$, $Z$, and $W$ are considered independent and identically distributed. In Scenario I, the random variables follow a uniform distribution on $\left[-\delta,\delta\right]$, and in Scenario II a standard normal distribution, ${\mathcal{N}}(0,1)$.

Given the extensive derivations needed to prove the results we have established, we only highlight in this section the less intuitive or most relevant aspects. The complete derivations can be found in \cite{Pascoal:2014}.

The theoretical evaluation  of the feature selection methods, whose objective functions are summarized in Table \ref{tab:feature_selection_methods}, need the following expressions:
\begin{itemize}
	\item[(i)] Entropy of the class, $C_k$, and of all features, $V_i$, $i=1,\ldots,10$ (see Tables \ref{tab:general_entropy_mi_true_scenario_1} and \ref{tab:general_entropy_mi_true_scenario_2} for Scenarios I and II, respectively).
	\item[(ii)] MI between the class and each feature, $\textrm{MI}(C_k,V_i)$, $i=1,\ldots,10$ (see also Tables \ref{tab:general_entropy_mi_true_scenario_1} and \ref{tab:general_entropy_mi_true_scenario_2} for Scenarios I and II, respectively).
	\item[(iii)]  MI between each pair of features, $\textrm{MI}(V_i,V_j)$, $i,j \in\{1,\ldots,10\}$ (see Table \ref{tab:mi_features_true_scenario_1} for Scenario I and Table \ref{tab:mi_features_true_scenario_2} for Scenario II).
\end{itemize}


From Table \ref{tab:general_entropy_mi_true_scenario_1}, we realise that if $X \sim \textrm{Unif}\left(-\delta,\delta\right)$, the entropy of $X$ is $h(X)=\ln(2\delta)$. This is a known result~\cite{Thomas:2006}. Note, however, that if $\delta=0.5$, $h(X)=0$ and if $0<\delta<0.5$ then $h(X)<0$, which stresses the fact that the entropies of continuous and discrete features do not have the same properties and require different interpretations. In next section we are going to show the impact of this fact in the performance of feature selection methods. Given its complexity, the general expression for $\textrm{MI}\left(C_k,X\right)$ is only defined for $\delta \geq 0.5$; its general form is provided in \cite{Pascoal:2014}.

With the exception of $Y^2 \stackrel{d}{=} X^2$ and $X-Y$, the derivation of the density functions of the features in Scenario I is quite simple. For the first case, one has
\begin{equation}\label{eq:fdp_f3_scenario_1}
\dps f_{Y^2}(u) = \left\{
\begin{array}{l l}
\dps \frac{1}{2\delta\sqrt{u}}, & \quad 0\leq u \leq \delta^2 \\[0.2cm]
0, & \quad \textrm{elsewhere}\\
\end{array} \right..
\end{equation}
As this is not a commonly known distribution, its entropy is calculated, leading to the expression provided in Table \ref{tab:general_entropy_mi_true_scenario_1}.
It can be shown that if $X$ and $Y$ are two independent features with $\rm{Unif}(-\delta,\delta)$ then $X-Y$ has a Triangular distribution with lower limit $-2\delta$, upper limit $2\delta$ and mode $0$, \ie, $X-Y \sim \textrm{Tri}\left(-2\delta,2\delta,0\right)$. The entropy of this triangular distribution is known and is provided in Table \ref{tab:general_entropy_mi_true_scenario_1}. For simplicity, in Table \ref{tab:general_entropy_mi_true_scenario_1} we present $\textrm{MI}\left(C_k,X-Y\right)$ only for the case when $\delta=0.5$, value suggested in \cite{Kwak:2002} and \cite{Huang:2008}; the general expression of this entropy is available in \cite{Pascoal:2014}.

A non-intuitive result, true for scenarios I and II, is that $\textrm{MI}\left(C_k,X^2\right)=0$ (proof provided in Appendix A), meaning that even though $X$ is important in the definition of $C_k$, $X^2$ has no association with the class.
Another relevant fact is that, for Scenario I, $\textrm{MI}\left(C_k,Y\right)$ does not depend on $\delta$.

In Scenario II, the features $X,\, Y,\, Z$ and $W$ have standard normal distribution, whose known good properties guarantee that all features under study have known distributions. Nevertheless, we raise the attention to $X^2$ (similarly $Y^2$ and $Z^2$) which have chi-squared distribution with one degree of freedom, $\chi^2_{(1)}$. It is known that the entropy of a random variable with chi-squared distribution with $k$ degrees of freedom is $k/2+\ln\left(2\Gamma\left(k/2\right)\right)+\left(1-k/2\right)\psi\left(k/2\right)$ where $\Gamma\left(\cdot\right)$ is the Gamma function and $\psi\left(\cdot\right)$ is the Digamma function~\cite{Lazo:1978}.

The calculation of MI between each feature and the class (apart from cases of independence) requires the use of the family of univariate skew normal distributions. This family generalizes the normal distribution, allowing skewness different from zero~\cite{Azzalini:1985}. A feature $V$ with skew normal distribution with location $\mu \in \Re$, scale $\sigma>0$, and shape $\alpha \in \mathbb{R}$ is represented by $\textrm{SN}(\mu,\sigma,\alpha)$. Note that, $(V-\mu)/\sigma \sim \textrm{SN}(0,1,\alpha)$, and if $\alpha=0$ then $V \sim\mathcal{N}(\mu,\sigma)$.

We recall that the MI is symmetric and non-negative (being $0$ for independent features), is invariant under scale or location transformations and under one-to-one transformations (that is, $\textrm{MI}(X,Y) = \textrm{MI}(U,W)$ where $u=g(x)$, $w=g(y)$, and $g$ is invertible, see \cite{malasianos:2010} for details). These properties justify the following results:
\begin{itemize}
	\item[(i)] All zeros in Tables \ref{tab:mi_features_true_scenario_1} and \ref{tab:mi_features_true_scenario_2};
	\item[(ii)]
	$\textrm{MI}\left(X,X-Y\right)$ = $\textrm{MI}\left(aX+b,X-Y\right)$\\ = $\textrm{MI}\left(Y,X-Y\right)$ = $\textrm{MI}\left(Z,W+Z\right)$\\ =
	$\textrm{MI}\left(W+d,W+Z\right)$, and
	\item[(iii)] $\textrm{MI}\left(X^2,X-Y\right)$=$\textrm{MI}\left(Y^2,X-Y\right)$=$\textrm{MI}\left(Z^2,W+Z\right)$.
\end{itemize}
Once again, it can be proved (vide \cite{Pascoal:2014} for details) that:
\begin{equation*}
\textrm{MI}\left(X,X-Y\right)=\left\{\begin{array}{ll}
\frac{1}{2},    & \rm{Scenario\; I}\\
\frac{\ln2}{2}, & \rm{Scenario\; II}
\end{array}
\right..
\end{equation*}
In a similar manner, it can be established that
%
%
\begin{flalign*}
\textrm{MI}\left(Y^2,X-Y\right)=\left\{\begin{array}{ll}
\frac{1-\ln2}{2},    & \rm{Scenario\; I}\\
-1+ \frac{\ln2}{2}+E\left(\ln{\left(\cosh{\left[(X-Y)|Y|\right]}\right)}\right), & \rm{Scenario\; II.}
\end{array}
\right.
\end{flalign*}

For Scenario II, the expression for $\textrm{MI}\left(Y^2,X-Y\right)$
is evaluated numerically, using the software  Mathematica~\cite{Mathematica:2003}, leading to the approximation $0.1078$, value included in Table \ref{tab:mi_features_true_scenario_2}.

Even though the MI between two identical discrete features is equal to the entropy of the feature, this property does not hold for absolute continuous features. In fact, \cite{Kotz:1966} proved that if $V_i$ and $V_j$ are two absolute continuous features, where $V_j$ is a measurable function of $V_i$, then $\textrm{MI}(V_i,V_j)=+\infty$.  This leads to:
\begin{itemize}
	\item[(i)] $\textrm{MI}\left(V_i,V_i\right)=+\infty$, for $i=1,\ldots,10$ and
	\item[(ii)] 
	%
	%
	\begin{flalign*}
	\textrm{MI}\left(X,aX+b\right)&=\textrm{MI}\left(aX+b,X^2\right)=\textrm{MI}\left(X,X^2\right)=\textrm{MI}\left(Y,Y^2\right)&\\
	&=\textrm{MI}\left(Z,Z^2\right)=+\infty.&
	\end{flalign*}
	
	%
\end{itemize}
All these results are summarized in Tables \ref{tab:mi_features_true_scenario_1} and \ref{tab:mi_features_true_scenario_2}.

\begin{table*}[!h]
	\centering
	\scriptsize
	\caption{General expression for the entropy of each feature, $h(V_i),i=1,\ldots,10$, and for the MI between each feature and the class, $\textrm{MI}\left(C_k,V_i\right),\, i=1,\ldots,10$, for Scenario I.}
	\begin{tabular}{lr@{$\sim$}lccc}
		\toprule
		Feature & \multicolumn{2}{c}{Description} & $h(\cdot)$ &  $\textrm{MI}\left(C_{k}, \cdot\right)$ \\
		\midrule
		$V_1$   & $X$ & $\textrm{Unif}\left(-\delta,\delta\right)$ & $\ln\left(2\delta\right)$ & $\dps -\frac{k}{2}+\ln\left(2\right)$ \\[0.3cm]  
		$V_2$   & $aX$+$b$ & $\textrm{Unif}\left(-\delta a+b,\delta a+b\right)$ & $\ln\left(2a\delta\right)$  &$\textrm{MI}\left(C_{k}, V_1\right)$  \\[0.3cm]
		$V_3$   &  \multicolumn{2}{c}{$\dps Y^2$}& $\ln\left(2\delta^2\right)-1$ & $0$  \\[0.3cm]
		$V_4$   & $X$-$Y$ & $\textrm{Tri}\left(-2\delta,2\delta,0\right)$ & $\dps \frac{1}{2}+\ln\left(2\delta\right)$ & $\dps \frac{-\left(k-1\right)^2\ln\left(1-k\right)}{4k}$\\[0.3cm] 
		$V_5$   & \multicolumn{2}{c}{$Z\stackrel{d}{=}X$} & $h(V_1)$ & $0$\\[0.3cm]
		$V_6$   & \multicolumn{2}{c}{$Z^2\stackrel{d}{=}Y^2$} & $h(V_3)$ & $0$\\[0.3cm]
		$V_7$   & \multicolumn{2}{c}{$Y\stackrel{d}{=}X$} & $h(V_1)$ & $\dps \frac{\left(k^2+1\right)\ln\left(\frac{1+k}{1-k}\right)+2k\left(\ln\left(1-k^2\right)-1\right)}{4k}$\\[0.3cm]
		$V_8$   & \multicolumn{2}{c}{$X^2\stackrel{d}{=}Y^2$} & $h(V_3)$ & $0$\\[0.3cm]
		$V_9$   &$W$+2& $\textrm{Unif}\left(-\delta+d,\delta+d\right)$ & $h(V_1)$ & $0$\\[0.3cm]
		$V_{10}$  & \multicolumn{2}{c}{$Z$+$W\stackrel{d}{=}X$-$Y$}& $h(V_4)$ & $0$  \\
		\bottomrule
	\end{tabular}
	\label{tab:general_entropy_mi_true_scenario_1}
\end{table*}

\begin{table*}[!h]
	\centering
	\scriptsize
	\caption{General expression for the entropy of each feature, $h(V_i),i=1,\ldots,10$, and for the MI between each feature and the class, $\textrm{MI}\left(C_k,V_i\right),\, i=1,\ldots,10$, for Scenario II. In $h(V_3)$, $\gamma$ represents the Euler's constant. }
	\begin{tabular}{lccc}
		\toprule
		Feature & Description & $h(\cdot)$ &  $\textrm{MI}\left(C_{k}, \cdot\right)$ \\
		\midrule
		\multirow{2}{*}{$V_1$}  & \multirow{2}{*}{$X\sim\mathcal{N}\left(0,1\right)$} & \multirow{2}{*}{$\dps \frac{1}{2}\ln(2 \pi e) $}  & $\dps \textrm{E}\left[\ln\left(2\Phi\left(-\frac{\textrm{A}_0}{k}\right) \Phi\left(\frac{\textrm{A}_1}{k}\right)\right)\right]$ where\\[0.3cm]
		&  &  & $\textrm{A}_0 \sim \textrm{SN}(0,1,-1/k)$ and $\textrm{A}_1 \sim \textrm{SN}(0,1,1/k)$\\[0.5cm]
		$V_2$   & $aX$+$b\sim\mathcal{N}\left(b,a^2\right)$ & $\dps \frac{1}{2}\ln\left(2 \pi e a^2\right)$& $\textrm{MI}\left(C_{k}, V_1\right)$\\[0.2cm]
		$V_3$   & $\dps Y^2\sim\chi^2_{\left(1\right)}$ & $\dps \frac{1}{2}(1+\ln \pi- \gamma)$ & $0$ \\[0.2cm] 
		\multirow{2}{*}{$V_4$} &  \multirow{2}{*}{$X-Y\sim\mathcal{N}\left(0,2\right)$}& \multirow{2}{*}{$\dps \frac{1}{2}\ln\left(4 \pi e\right)$} & $\dps \textrm{E}\left[\ln\left(2\Phi\left(-\frac{1-k}{k+1} \textrm{B}_0\right) \Phi\left(\frac{1-k}{k+1}\textrm{B}_1\right)\right)\right]$ where\\[0.3cm]
		&  &  & \scriptsize{$\textrm{B}_0 \sim \textrm{SN}\left(0,1,-\dps \frac{1-k}{k+1}\right)$ and $\textrm{B}_1 \sim \textrm{SN}\left(0,1,\dps \frac{1-k}{k+1}\right)$} \\[0.5cm]
		$V_5$   & $Y\stackrel{d}{=}X$ & $h(V_1)$ & $0$\\[0.2cm]
		$V_6$   & $Z^2\stackrel{d}{=}Y^2$ & $h(V_3)$ & $0$\\[0.2cm]
		\multirow{2}{*}{$V_7$} &  \multirow{2}{*}{$Y\stackrel{d}{=}X$} & \multirow{2}{*}{$h(V_1)$} & $\dps \textrm{E}\left[\ln\left(2\Phi(-k \textrm{D}_0) \Phi(k\textrm{D}_1)\right)\right]$ where\\[0.3cm]
		& &  & $\textrm{D}_0 \sim \textrm{SN}\left(0,1,-k\right)$ and $\textrm{D}_1 \sim \textrm{SN}\left(0,1,k\right)$\\[0.5cm]
		$V_8$   & $X^2\stackrel{d}{=}Y^2$ & $h(V_3)$ & $0$\\[0.2cm]
		$V_9$      & $W$+$d\sim\mathcal{N}\left(d,1\right)$&   $h(V_1)$ & $0$\\[0.2cm]
		$V_{10}$   &$W$+$Z\sim\mathcal{N}\left(0,2\right)$ & $h(V_4)$ & $0$\\
		\bottomrule
	\end{tabular}
	\label{tab:general_entropy_mi_true_scenario_2}
\end{table*}

\begin{table*}[!h]
	\centering
	\scriptsize
	\caption{MI between the input features, $\textrm{MI}\left(V_i,V_j\right), i=1,\ldots,10, \, j=1,\ldots,i$, for Scenario I.}
	\begin{tabular}{c m{0.7cm}m{0.7cm}m{0.7cm}m{0.7cm}m{0.7cm}m{0.7cm}m{0.7cm}m{0.7cm}m{0.7cm}m{1cm}}
		\toprule
		$\textrm{MI}\left(\cdot,\cdot\right)$ & $V_1$ & $V_2$ & $V_3$ & $V_4$ & $V_5$ & $V_6$ & $V_7$ & $V_8$ & $V_9$ & $V_{10}$\\
		\midrule
		$V_1$    & $+\infty$  & & & & & & & & &\\
		$V_2$    & $+\infty$  & $+\infty$  & & & & & & & &\\
		$V_3$    &   $0$      &    $0$     & $+\infty$  & & & & & & &\\
		$V_4$    &  $0.5$     &   $0.5$    & $\frac{1-\ln2}{2}$ & $+\infty$  & & & & & &\\
		$V_5$    &   $0$      &    $0$     &    $0$     &    $0$     &  $+\infty$ & & & & &\\
		$V_6$    &   $0$      &    $0$     &    $0$     &    $0$     &  $+\infty$ & $+\infty$ & & & &\\
		$V_7$    &   $0$      &    $0$     & $+\infty$  &   $0.5$    &    $0$     &    $0$    & $+\infty$ & & &\\
		$V_8$    & $+\infty$  & $+\infty$  &    $0$     & $\frac{1-\ln2}{2}$   &    $0$     &    $0$    &    $0$    & $+\infty$ & &\\
		$V_9$    &   $0$      &    $0$     &    $0$     &    $0$     &    $0$     &    $0$    &    $0$    &     $0$    & $+\infty$ & \\
		$V_{10}$ &   $0$      &    $0$     &    $0$     &    $0$     &   $0.5$    &  $\frac{1-\ln2}{2}$ &    $0$    &     $0$    &    $0.5$  & $+\infty$\\
		\bottomrule
	\end{tabular}
	\label{tab:mi_features_true_scenario_1}
\end{table*}

\begin{table*}[!h]
	\centering
	\scriptsize
	\caption{MI between the input features, $\textrm{MI}\left(V_i,V_j\right),\, i=1,\ldots,10, \, j=1,\ldots,i$, for Scenario II.}
	\begin{tabular}
		{c m{0.7cm}m{0.7cm}m{0.7cm}m{0.7cm}m{0.7cm}m{0.7cm}m{0.7cm}m{0.7cm}m{0.7cm}m{1cm}}
		\toprule
		$\textrm{MI}\left(\cdot,\cdot\right)$ & $V_1$ & $V_2$ & $V_3$ & $V_4$ & $V_5$ & $V_6$ & $V_7$ & $V_8$ & $V_9$ & $V_{10}$\\
		\midrule
		$V_1$    & $+\infty$  & & & & & & & & & \\
		$V_2$    & $+\infty$  & $+\infty$  & & & & & & & & \\
		$V_3$    &   $0$      &    $0$     & $+\infty$  & & & & & & & \\
		$V_4$    & $\frac{\ln2}{2}$ & $\frac{\ln2}{2}$ & $0.1078$ & $+\infty$  & & & & & & \\
		$V_5$    &   $0$      &    $0$     &    $0$     &    $0$   &  $+\infty$ & & &  & & \\
		$V_6$    &   $0$      &    $0$     &    $0$     &    $0$   &  $+\infty$ & $+\infty$ & & & & \\
		$V_7$    &   $0$      &    $0$     & $+\infty$  & $\frac{\ln2}{2}$ &    $0$     &    $0$    & $+\infty$ &  & & \\
		$V_8$    & $+\infty$  & $+\infty$  &    $0$     & $0.1078$ &    $0$     &    $0$    &    $0$    & $+\infty$ & & \\
		$V_9$    &   $0$      &    $0$     &    $0$     &    $0$   &    $0$     &    $0$    &    $0$    &    $0$    & $+\infty$ & \\
		$V_{10}$ &   $0$      &    $0$     &    $0$     &    $0$   &  $\frac{\ln2}{2}$  & $0.1078$  &    $0$    &    $0$    & $\frac{\ln2}{2}$  & $+\infty$ \\
		\bottomrule
	\end{tabular}
	\label{tab:mi_features_true_scenario_2}
\end{table*}

\section{Theoretical feature ordering}\label{sec:theoretical_feature_ordering}

\noindent In this section, we will present the true feature ordering obtained with the evaluation scenario described in Section \ref{sec:eval:scen}, and using the results of Section \ref{sec:theoretical_h_mi}. As in \cite{Huang:2008}, we consider for feature $aX+b$ that $a=3$ and $b=1$ and that the uniform distribution has parameter $\delta=0.5$; we also consider that $d=2$. These concretizations lead to the entropy and MI (with the class) values shown in Table \ref{tab:entropy_mi_true_scenario_1} (for scenarios I and II).

\begin{table*}[!h]
	\caption{Entropy of each input variable, $h(V_i),i=1,\ldots,10$, and MI between each input variable and the class variable, $\textrm{MI}(C_k,V_i), \,k=0.2,0.8,\, i=1,\ldots,10$, for Scenario I and II, $a=3,\, b=1$, $d=2$, and $\delta=0.5$.}
	\label{tab:entropy_mi_true_scenario_1}
	\centering
	\scriptsize 
	\begin{tabular}{c||c|ccc||c|ccc}
		\toprule
		& \multicolumn{4}{c||}{Scenario I} & \multicolumn{4}{c}{Scenario II}\\ 
		\cline{2-5}\cline{6-9}
		Feature & 
		Dist. & 
		$h(\cdot)$ &  
		$\textrm{MI}(\cdot,C_{0.2})$ & 
		$\textrm{MI}(\cdot,C_{0.8})$ & 
		Dist. & 
		$h(\cdot)$ &  
		$\textrm{MI}(\cdot,C_{0.2})$ & 
		$\textrm{MI}(\cdot,C_{0.8})$ \\
		\midrule
		$V_1$	&	$\textrm{Unif}(-\frac{1}{2},\frac{1}{2})$	&	$0$	&	$0.5932$	&	$0.2932$	& $\mathcal{N}(0,1)$	&	$1.4189$	&	$0.5520$	&	$0.2495$			\\[0.3cm]
		$V_2$	&	$\textrm{Unif}(-\frac{1}{2},\frac{5}{2})$	&	$1.0986$	&	$0.5932$	&	$0.2932$	& $\mathcal{N}(1,3^2)$	&	$2.5176$	&	$0.5520$	&	$0.2495$			\\[0.3cm]
		$V_3$	&	$\dps Y^2$	&	$-1.6932$	&	$0$	&	$0$	& $\dps \chi^2_{(1)}$	&	$0.7838$	&	$0$	&	$0$			\\[0.3cm]
		$V_4$	&	$\textrm{Tri}(-1,1,0)$	&	$0.5000$	&	$0.1785$	&	$0.0201$	& $\mathcal{N}(0,2)$&$1.7655$&$0.0947$	&	$0.0032$							\\[0.3cm]
		$V_5$	&	$\textrm{Unif}(-\frac{1}{2},\frac{1}{2})$	&	$0$	&	$0$	&	$0$	&	$\mathcal{N}(0,1)$	&	$1.4189$	&	$0$	&	$0$		\\[0.3cm]
		$V_6$	&	$Z^2\stackrel{d}{=}V_3$	&	$-1.6932$	&	$0$	&	$0$	& $\dps \chi^2_{(1)}$	&	$0.7838$	&	$0$	&	$0$			\\[0.3cm]
		$V_7$	&$\textrm{Unif}(-\frac{1}{2},\frac{1}{2})$	&	$0$	&	$0.0067$	&	$0.1153$	& $\mathcal{N}(0,1)$	&	$1.4189$&$0.0124$&	$0.1434$						\\[0.3cm]
		$V_8$	&	$X^2\stackrel{d}{=}V_3$	&	$-1.6932$	&	$0$	&	$0$	& $\dps \chi^2_{(1)}$	&	$0.7838$	&	$0$	&	$0$			\\[0.3cm]
		$V_9$	&	$\textrm{Unif}(\frac{3}{2},\frac{5}{2})$	&	$0$	&	$0$	&	$0$	&$\mathcal{N}(2,1)$	&	$1.4189$	&$0$	&$0$					\\[0.3cm]
		$V_{10}$	&		$\textrm{Tri}(-1,1,0)$	&	$0.5000$	&	$0$	&	$0$	&$\mathcal{N}(0,2)$	&	$1.7655$	&	$0$	&	$0$			\\
		\bottomrule
	\end{tabular}
\end{table*}

The feature ordering results are shown in Table \ref{tab:order_true_scenario_1_k02} (for Scenario I) and in Table \ref{tab:order_true_scenario_2_k02} (for Scenario II).

\begin{table*}[!h]
	\centering
	\scriptsize
	\caption{Scenario I - Feature selection ordering, (a) $k=0.2$ and (b) $k=0.8$. The methods for which the two first selected features form a relevance-optimal set are shown in bold type.}
	\begin{tabular}{l cccccccccc}
		\toprule
		Methods & \multicolumn{10}{c}{Order of feature selection}\\
		\midrule
		\textbf{MIFS} ($\beta=0$) & $X$&$X$-$Y$&$Y$&$Z$&$W$+2&$Z$+$W$&-&-&-&-\\[0.1cm]
		\textbf{MIFS} ($\beta=.4,.7,1$) & $X$&$Y$&$Z$&$W$+2&$X$-$Y$&$Z$+$W$&$3$$X$+$1$&$Y^2$&$Z^2$&$X^2$\\[0.1cm]
		MIFS-U ($\beta=0$) & $X$&-&-&-&-&-&-&-&-&-\\[0.1cm]
		MIFS-U ($\beta=.4,.7,1$) & $X$&$3$$X$+$1$&$X$-$Y$&$X^2$&-&-&-&-&-&-\\[0.1cm]
		\textbf{mRMR} & $X$&$Y$&$Z$&$W$+2&$X$-$Y$&$Z$+$W$&$3$$X$+$1$&$Y^2$&$Z^2$&$X^2$\\[0.1cm]
		mMIFS-U & $X$&$3$$X$+$1$&$X$-$Y$&$X^2$&-&-&-&-&-&-\\ [0.1cm]
		MICC & $X$&$X^2$&$X$-$Y$&$Y^2$&-&-&-&-&-&-\\[0.1cm]
		QMIFS & $X$&$3$$X$+$1$&-&-&-&-&-&-&-&-\\[0.1cm]
		NMIFS & $X$&$X^2$&$Y^2$&$Z^2$&$X$-$Y$&-&-&-&-&-\\[0.1cm]
		\textbf{maxMIFS} & $X$&$Y$&$Z$&$W$+2&$X$-$Y$&$Z$+$W$&$3$$X$+$1$&$Y^2$&$Z^2$&$X^2$\\
		\bottomrule
	\end{tabular}
	\begin{center}
		\vspace*{-2mm}
		(a) $\quad k=0.2.$\\
	\end{center}
	\centering
	\scriptsize
	\begin{tabular}{l cccccccccc}
		\toprule
		Methods & \multicolumn{10}{c}{Order of feature selection}\\
		\midrule
		\textbf{MIFS} ($\beta=0$) & $X$&$Y$&$X$-$Y$&$Z$&$W$+2&$Z$+$W$&-&-&-&-\\[0.1cm]
		\textbf{MIFS} ($\beta=.4,.7,1$) & $X$&$Y$&$Z$&$W$+2&$X$-$Y$&$Z$+$W$&$3$$X$+$1$&$Y^2$&$Z^2$&$X^2$\\[0.1cm]
		MIFS-U ($\beta=0$) & $X$&-&-&-&-&-&-&-&-&-\\[0.1cm]
		MIFS-U ($\beta=.4,.7,1$) & $X$&$3$$X$+$1$&$X$-$Y$&$X^2$&-&-&-&-&-&-\\[0.1cm]
		\textbf{mRMR} & $X$&$Y$&$Z$&$W$+2&$X$-$Y$&$Z$+$W$&$3$$X$+$1$&$Y^2$&$Z^2$&$X^2$\\[0.1cm]
		mMIFS-U & $X$&$3$$X$+$1$&$X$-$Y$&$X^2$&-&-&-&-&-&-\\ [0.1cm]
		MICC & $X$&$X^2$&$X$-$Y$&$Y^2$&-&-&-&-&-&-\\[0.1cm]
		QMIFS & $X$&$3$$X$+$1$&-&-&-&-&-&-&-&-\\[0.1cm]
		NMIFS & $X$&$X^2$&$3$$X$+$1$&$Z^2$&$X$-$Y$&-&-&-&-&-\\[0.1cm]
		\textbf{maxMIFS} & $X$&$Y$&$Z$&$W$+2&$X$-$Y$&$Z$+$W$&$3$$X$+$1$&$Y^2$&$Z^2$&$X^2$\\
		\bottomrule
	\end{tabular}
	\begin{center}
		\vspace*{-2mm}
		(b) $\quad k=0.8.$\\
	\end{center}
	\vspace*{-5mm}
	\label{tab:order_true_scenario_1_k02}
\end{table*}

\begin{table*}[!h]
	\centering
	\scriptsize
	\caption{Scenario II - Feature selection ordering, (a) $k=0.2$ and (b) $k=0.8$. The methods for which the two first selected features form a relevance-optimal set are shown in bold type.}
	\begin{tabular}{l cccccccccc}
		\toprule
		Methods & \multicolumn{10}{c}{Order of feature selection}\\
		\midrule
		\textbf{MIFS} ($\beta=0$) & $X$&$X$-$Y$&$Y$&$Z$&$W$+2&$Z$+$W$&-&-&-&-\\[0.1cm]
		\textbf{MIFS} ($\beta=.4,.7,1$) & $X$&$Y$&$Z$&$W$+2&$X$-$Y$&$Z$+$W$&$3$$X$+$1$&$Y^2$&$Z^2$&$X^2$\\[0.1cm]
		\textbf{MIFS-U} ($\beta=0$) & $X$&$X$-$Y$&$Y$&$Z$&$W$+2&$Z$+$W$&-&-&-&-\\[0.1cm]
		\textbf{MIFS-U} ($\beta=.4$) & $X$&$X$-$Y$&$Y$&$Z$&$W$+2&$Z$+$W$&$3$$X$+$1$&$Y^2$&$X^2$&-\\[0.1cm]
		\textbf{MIFS-U} ($\beta=.7,1$) & $X$&$Y$&$Z$&$W$+2&$Z$+$W$&$X$-$Y$&$3$$X$+$1$&$Y^2$&$X^2$&-\\[0.1cm]
		\textbf{mRMR} & $X$&$Y$&$Z$&$W$+2&$X$-$Y$&$Z$+$W$&$3$$X$+$1$&$Y^2$&$Z^2$&$X^2$\\[0.1cm]
		\textbf{mMIFS-U} & $X$&$Y$&$Z$&$W$+2&$Z$+$W$&$X$-$Y$&$3$$X$+$1$&$Y^2$&$X^2$&-\\ [0.1cm]
		\textbf{MICC} & $X$&$Y$&$X$-$Y$&$Y^2$&$X^2$&$3$$X$+$1$&-&-&-&-\\[0.1cm]
		\textbf{QMIFS} & $X$&$Y$&$Z$&$W$+2&$Z$+$W$&$X$-$Y$&-&-&-&-\\[0.1cm]
		\textbf{NMIFS} & $X$&$Y$&$Z$&$W$+2&$X$-$Y$&$Z$+$W$&$3$$X$+$1$&$Y^2$&$Z^2$&$X^2$\\[0.1cm]
		\textbf{maxMIFS} & $X$&$Y$&$Z$&$W$+2&$X$-$Y$&$Z$+$W$&$3$$X$+$1$&$Y^2$&$Z^2$&$X^2$\\
		\bottomrule
	\end{tabular}
	\begin{center}
		\vspace*{-2mm}
		(a) $\quad k=0.2.$\\
	\end{center}
	\centering
	\scriptsize
	\begin{tabular}{l cccccccccc}
		\toprule
		Methods & \multicolumn{10}{c}{Order of feature selection}\\
		\midrule
		\textbf{MIFS} ($\beta=0$) & ${X}$&$Y$&$X$-$Y$&$Z$&$W$+2&$Z$+$W$&-&-&-&-\\[0.1cm]
		\textbf{MIFS} ($\beta=.4,.7,1$) & ${X}$&$Y$&$Z$&$W$+2&$X$-$Y$&$Z$+$W$&$3$$X$+$1$&$Y^2$&$Z^2$&$X^2$\\[0.1cm]
		\textbf{MIFS-U} ($\beta=0$) & $X$&$Y$&$X$-$Y$&$Z$&$W$+2&$Z$+$W$&-&-&-&-\\[0.1cm]
		\textbf{MIFS-U} ($\beta=.4,.7,1$) & $X$&$Y$&$Z$&$W$+2&$Z$+$W$&$X$-$Y$&$3$$X$+$1$&$Y^2$&$X^2$&-\\[0.1cm]
		
		\textbf{mRMR} & $X$&$Y$&$Z$&$W$+2&$X$-$Y$&$Z$+$W$&$3$$X$+$1$&$Y^2$&$Z^2$&$X^2$\\[0.1cm]
		\textbf{mMIFS-U} & $X$&$Y$&$Z$&$W$+2&$Z$+$W$&$X$-$Y$&$3$$X$+$1$&$Y^2$&$X^2$&-\\ [0.1cm]
		\textbf{MICC} & $X$&$Y$&$X$-$Y$&$Y^2$&$X^2$&$3$$X$+$1$&-&-&-&-\\[0.1cm]
		\textbf{QMIFS} & $X$&$Y$&$Z$&$W$+2&$Z$+$W$&$X$-$Y$&-&-&-&-\\[0.1cm]
		\textbf{NMIFS} & $X$&$Y$&$Z$&$W$+2&$X$-$Y$&$Z$+$W$&$3$$X$+$1$&$Y^2$&$Z^2$&$X^2$\\[0.1cm]
		\textbf{maxMIFS} & $X$&$Y$&$Z$&$W$+2&$X$-$Y$&$Z$+$W$&$3$$X$+$1$&$Y^2$&$Z^2$&$X^2$\\
		\bottomrule
	\end{tabular}
	\begin{center}
		\vspace*{-2mm}
		(b) $\quad k=0.8.$\\
	\end{center}
	\vspace*{-5mm}
	\label{tab:order_true_scenario_2_k02}
\end{table*}

We have seen in previous sections that features can have null and negative entropy, and null MI with the class. It was also seen that the MI between features can be null or $+\infty$. These limiting values have a deep impact in the performance of  feature selection methods, a characteristic that seems not have been accounted for in previous works.

From Tables \ref{tab:order_true_scenario_1_k02} and \ref{tab:order_true_scenario_2_k02} it is clear that, in many cases, it was not possible to determine an ordering for all ten features. There were only three methods for which this was always possible: MIFS ($\beta \neq 0$), mRMR, and maxMIFS.

These methods achieved the same ordering in all situations: $X$, $Y$, $Z$, $W+2$, $X-Y$, $Z+W$, $3X+1$, $Y^2$, $Z^2$, and $X^2$. To understand why, consider Scenario I and the MIFS method (with $\beta=1$), which is probably the easiest to follow.
\begin{itemize}
	\item $X$ is selected first because, together with $3X+1$ has the highest MI with the class, both for $k=0.2$ and $k=0.8$. $X$ is selected before $3X+1$ simply because it was placed first in the list of initial features. As a result, $3X+1$ becomes redundant and will only be selected in seventh place.
	\item $Y$ is selected in second place, because it is informative about the class and is not redundant with $X$. One may think that $X-Y$ should be selected at this step, because $(X,X-Y)$ is a relevance-optimal set and its MI with the class is larger than that of $Y$, especially when $k=0.2$. However, $X-Y$ is also more redundant with $X$. Indeed, for candidate $Y$, the objective function value is simply $\textrm{MI}\left(C_{0.2},Y\right)=0.0067$ while, for $X-Y$, is -0.3215, since $\textrm{MI}\left(C_{0.2},X-Y\right)=0.1785$ and $\textrm{MI}\left(X,X-Y\right) = 0.5$.
	\item The next two features to be selected are $Z$ and $W+2$, both with a null objective function. Note that, after selecting $Y$, the redundancy of $X-Y$ increases to $\textrm{MI}\left(X,X-Y\right)+\textrm{MI}\left(Y,X-Y\right) = 1$, reinforcing the negative value of its objective function. While selecting $Z$, features $Z^2$, $W+2$, and $Z+W$ have the same objective function value; they are left behind just because they are placed after $Z$ in the list of initial features.
	\item At the fifth step, the competition is between $X-Y$ and $Z+W$, because the remaining features, $3X+1$, $Y^2$, $Z^2$, and $X^2$, all have a $-\infty$ objective function value, since they are fully associated with at least one of the already selected features. $X-Y$ is selected in fifth place and $Z+W$ in sixth, because the former has some association with the class, the latter has not, and they both have the same redundancy.
	\item Finally, the last selected features, from seventh to tenth place, are $3X+1$, $Y^2$, $Z^2$, and $X^2$.
\end{itemize}

As expected, the best methods select first one relevance-optimal set, indeed one of the sets involving independent features. $(X,Y)$ was the chosen one but, depending on the order of features in the initial list, $(3X+1,Y)$ could also have been selected first. It is worth noting that the next two selected features, $Z$ and $W+2$, are irrelevant. We argue that this is a good characteristic of the methods given that, in practice, the number of optimal features is unknown and some features beyond the relevant ones may be selected. In general, it is less harmful for the classifier to select irrelevant features. For example, many classifiers require the inversion of the covariance matrix; and while the covariance matrix of $(X,Y,Z)$ is invertible, the one of $(X,Y,X-Y)$ is not.

The MIFS ($\beta \neq 0$), mRMR, and maxMIFS feature selection methods do not suffer from any kind of indeterminations, nor from the possibility of having negative entropies. This cannot be said about the other methods. In the following we give some examples.

Negative entropies can affect several of the proposed methods. When using NMIFS in Scenario I, both with $k=0.2$ and $k=0.8$, $X^2$ is selected in second place, because the design of the objective function ignored the possibility of a negative entropy. In this case, since the entropy of $X^2$ is negative, $\textrm{NI}\left(X^2,X\right)$ is also negative, which leads to a redundancy contribution that increases the objective function value, something certainly opposite to what the authors have wished. Another problem with negative entropies is that they can lead to indeterminations of the type $+\infty-\infty$. The methods that can be impaired by negative entropies are MIFS-U, mMIFS-U, MICC, QMIFS, and NMIFS.

MIFS $(\beta = 0)$ and MIFS-U $(\beta = 0)$ always select the features according to the MI between the features and the class, because there is no redundancy component involved in the objective function. For example, in the case of $k=0.2$ (both scenarios), the order is $X$, $3X+1$, $X-Y$, $Y$, $Y^2$, $Z$, $Z^2$, $X^2$, $W+2$, and $Z+W$, according to the values of Table \ref{tab:entropy_mi_true_scenario_1}. However, if the implementation of the algorithm does not ignore the redundancy component and considers $\beta$ explicitly, many of the features cannot be selected due to indeterminations of the type $0 \times (+\infty)$. In the cases of MIFS $(\beta = 0)$ (both scenarios) and MIFS-U $(\beta = 0)$ (Scenario II), only six features can be selected. In Scenario I, MIFS-U behaves even worse. Because the objective function includes $\beta$ in the numerator and the entropy of an already selected feature, $h(V_s)$, in the denominator, and since the first selected feature, $X$, has null entropy, there will always be an indetermination of type $0/0$, and no additional feature can be selected. This shows that special care must be exercised when using MIFS and MIFS-U with $\beta = 0$.

MIFS-U $(\beta=0.4,0.7,1)$ and mMIFS-U select as second, third, and fourth features, $3X+1$, $X-Y$, and $X^2$, and from then on are unable to select other features, in Scenario I. The features $3X+1$, $X-Y$, and $X^2$ are all selected with a $-\infty$ objective function value; they appear in this order because it is how they appear in the initial list of features. The remaining selection steps are impaired by $0/0$ indeterminations. This is due to the quotient $\textrm{MI}(V_i,V_s)/h(V_s)$: since the entropy of $X$ is null, this quotient is either an indetermination, for independent features, or $+\infty$, for fully associated features. Thus, the algorithm selects only features fully associated with $X$, resulting in a quite wrong selection. As in the case of negative entropies, the methods that are impaired by $0/0$ indetermination are MIFS-U, mMIFS-U, MICC, QMIFS, and NMIFS.

The results obtained with MICC and QMIFS in Scenario I are affected by several types of indeterminations simultaneously. For example, QMIFS selects the second feature in the same way as MIFS-U $(\beta=0.4,0.7,1)$ and mMIFS-U, i.e., with a $-\infty$ objective function value. However, when trying to select the third feature, the objective function for candidate $Y^2$ includes $0/0$ and $0 \times \infty$ indeterminations, for candidate $X-Y$ includes $+\infty -\infty$, and so on.


When comparing Tables \ref{tab:order_true_scenario_1_k02} and \ref{tab:order_true_scenario_2_k02}, it is clear that there are fewer problems in Scenario II, i.e., when the random variables have a normal distribution. This is due to the fact that, unlike Scenario I, in Scenario II there are no features with null or negative entropies (Table \ref{tab:entropy_mi_true_scenario_1}). However, this is not a general property. For this set of features it would be possible to find negative entropies, if $\sigma^2<(2\pi e)^{-1}$, or null entropies, if $\sigma^2=(2\pi e)^{-1}$. Despite the improved behavior, in Scenario II it is not possible to terminate the selection process in MIFS $(\beta=0)$, MIFS-U, mMIFS-U, MICC, and QMIFS methods.

The results of MIFS-U, in Scenario II with $k=0.2$, illustrate an interesting issue regarding the weight given to the redundancy component. With $\beta = 0.4$, $X-Y$ is selected in second place and $Y$ in third, while with $\beta = 0.7$ and $\beta = 1$ the opposite occurs. Indeed, the association with the class is higher for $X-Y$ than for $Y$. But $Y$ has no redundancy with $X$ while $X-Y$ has some. For $\beta = 0.4$, the redundancy component is given relatively low weight, not enough for cancelling out the strength of the relevancy component, and $X-Y$ gets selected before $Y$.

A general conclusion can be drawn from this study. From the analysis of the objective functions and from the results of Tables \ref{tab:order_true_scenario_1_k02} and \ref{tab:order_true_scenario_2_k02}, there are only three methods that do not have problems with indeterminations, which are MIFS $(\beta\neq0)$, mRMR, and maxMIFS. However, due to the problems with MIFS $(\beta=0)$, we cannot recommend its use with small $\beta$ values. In all these three methods, the same order was obtained in all cases. The WR independent features, $X$ and $Y$ are selected first, then the irrelevant ones, $Z$ and $W+2$, and finally features that have become redundant with already selected ones.

\section{Simulation study}\label{sec:simulation}

\noindent To assess the performance of the eight feature selection methods (MIFS, MIFS-U, mRMR, mMIFS-U, QMIFS, MICC, NMIFS, and maxMIFS), as well as the importance of correctly estimating the MI, a simulation study based on the two evaluation scenarios presented in Section \ref{sec:eval:scen} was developed. We randomly generated $5000$ samples of sizes $n=50,\, 100,\, 500,\, 1000$, and 5000, for the ten input features shown in Table \ref{tab:input_features}, and applied the eight feature selection methods to each sample.


In Table \ref{tab:mean_entropy_MI} we compare our MI estimates with those obtained in \cite{Kwak:2002} and \cite{Huang:2008}, and with the true value obtained with the results of Section \ref{sec:theoretical_h_mi}. The case considered in \cite{Kwak:2002} and \cite{Huang:2008} was that of Scenario I with $k=0.2$, and a sample size $n=1000$. Our MI estimates were obtained indirectly using property (e) of Section \ref{sec:entropy:MI}, for $\hat{\textrm{MI}}\left(C_{0.2},V_i\right)$, and equation (d) of the same section, for ${\hat{\textrm{MI}}}\left(V_j,V_i\right)$. The differential  entropies were estimated by partitioning the simulated values in equal-width bins. We added the correction factor $\ln(\Delta)$ to the entropies estimated based on the discretized values, where $\Delta$ is the length of the bins (vide \cite{Thomas:2006,Pascoal:2014} for details). The number of bins considered is a function of the sample size given by $m=\lceil\sqrt{n}\rceil$.
As in \cite{Kwak:2002} and \cite{Huang:2008}, we present the mean of the MI estimates between the class and the first four features, $\hat{\textrm{MI}}\left(C_{0.2},V_i\right), i=1,2,3,4$, the mean of the MI estimates between the first feature, and the third and fourth features, $\hat{\textrm{MI}}\left(V_1,V_i\right), i=3,4$. The true values were extracted from Table \ref{tab:entropy_mi_true_scenario_1}. It is clear from Table \ref{tab:mean_entropy_MI} that the results of \cite{Kwak:2002} and \cite{Huang:2008} have large deviations relative to the true values. Our own estimates are much closer. We guess that these errors may be attributed to the fact that \cite{Kwak:2002}  and \cite{Huang:2008} only considered one sample, which is clearly insufficient for statistical confidence. Another possibility is that they either have used the same bin width in the discretization of univariate and bivariate distributions, or have not included the corresponding correction factor. We believe that estimation errors could be one reason explaining the quality overstatement of some proposed methods. In \cite{Pascoal:2012}, we have extended this study to other entropy and MI estimates for which we have estimated the mean square error, $\textrm{MSE}(\hat{h}(V_i)),\, i=1, \ldots,8$, and $\textrm{MSE}(\hat{\textrm{MI}}\left(C_k,V_i\right)),\, k=0.2, 0.8,\, i=1, \ldots, 8$, for scenarios I and II. In all cases the estimated values are close to the true ones.
\begin{table*}[!h]
	\centering
	\scriptsize
	\caption{Estimated (mean of the) MI between the class, $C_{0.2}$, and the first four features, $V_1$, $V_2$, $V_3$ and $V_4$, $\hat{\textrm{MI}}\left(C_{0.2},V_i\right), i=1,2,3,4$, and (mean of the) MI between the first feature, $V_1$, and the third and fourth features, $V_3$ and $V_4$, $\hat{\textrm{MI}}\left(V_1,V_i\right), i=3,4$, considering sample(s) of size $n=1000$.}
	\begin{tabular}{c cccccc}
		\toprule
		Source & $\hat{\textrm{MI}}\left(C_{0.2},V_i\right), i=1,2$ & $\hat{\textrm{MI}}\left(C_{0.2},V_3\right)$ & $\hat{\textrm{MI}}\left(C_{0.2},V_4\right)$  & $\hat{\textrm{MI}}\left(V_1,V_3\right)$ & $\hat{\textrm{MI}}\left(V_1,V_4\right)$  \\
		\midrule
		\cite{Kwak:2002} & $0.8459$ & $0.0170$ & $0.2621$ & $0.0610$ & $0.6168$  \\
		\cite{Huang:2008} & $0.8438$ &  $0.0383$ & $0.2807$ & $0.0634$ & $0.6099$ & \\
		Our estimate & $0.5932$ & $0.0075$ & $0.1779$ & $0.0107$ & $0.5004$   \\
		True & $0.5932$ & $0$ & $0.1785$ & $0$ & $0.5$ \\
		\bottomrule
	\end{tabular}
	\label{tab:mean_entropy_MI}
\end{table*}

\begin{table}[!h]
	\centering
	\scriptsize
	\caption{Relative frequency of the optimal subsets.}
	\begin{tabular}{lcccc}
		\toprule
		\multirow{2}{*}{Methods} & \multicolumn{2}{c}{Scenario I} & \multicolumn{2}{c}{Scenario II} \\
		\cmidrule(r){2-5}
		&    $k=0.2$    & $k=0.8$          &    $k=0.2$    & $k=0.8$ \\
		\midrule
		\multicolumn{1}{l|}{MIFS ($\beta=0$)} & \multicolumn{1}{c}{\hspace*{-6mm}0}   &    \multicolumn{1}{c|}{\hspace*{-6mm}0} & \multicolumn{1}{c}{\hspace*{-6mm}0} &\multicolumn{1}{c}{\hspace*{-8mm}0} \\[0.1cm]
		\multicolumn{1}{l|}{MIFS ($\beta=0.4$)} & 0.9744 & \multicolumn{1}{c|}{\hspace*{-6mm}1} & 0.9998 & \multicolumn{1}{c}{\hspace*{-8mm}1} \\[0.1cm]
		\multicolumn{1}{l|}{MIFS ($\beta=0.7$)} & 0.9502 & \multicolumn{1}{c|}{\hspace*{-6mm}1} & 0.9982 & \multicolumn{1}{c}{\hspace*{-8mm}1} \\[0.1cm]
		\multicolumn{1}{l|}{MIFS ($\beta=1$)} & 0.9310 & \multicolumn{1}{c|}{\hspace*{-6mm}1} & 0.9936 & \multicolumn{1}{c}{\hspace*{-8mm}1} \\[0.1cm]
		\multicolumn{1}{l|}{MIFS-U ($\beta=0$)} & \multicolumn{1}{c}{\hspace*{-6mm}0} & \multicolumn{1}{c|}{\hspace*{-6mm}0}&\multicolumn{1}{c}{\hspace*{-6mm}0} &\multicolumn{1}{c}{\hspace*{-8mm}0} \\[0.1cm]
		\multicolumn{1}{l|}{MIFS-U ($\beta=0.4$)} & \multicolumn{1}{c}{\hspace*{-6mm}0} & \multicolumn{1}{c|}{0.4024} & \multicolumn{1}{c}{\hspace*{-6mm}0} & 0.6720 \\[0.1cm]
		\multicolumn{1}{l|}{MIFS-U ($\beta=0.7$)}&0.3950 & \multicolumn{1}{c|}{0.4024} & 0.6556 & \multicolumn{1}{c}{\hspace*{-8mm}1}\\[0.1cm]
		\multicolumn{1}{l|}{MIFS-U ($\beta=1$)} & 0.3964 & \multicolumn{1}{c|}{0.4024} & 0.6556 & \multicolumn{1}{c}{\hspace*{-8mm}1}\\[0.1cm]
		\multicolumn{1}{l|}{mRMR} & 0.9310  & \multicolumn{1}{c|}{\hspace*{-6mm}1} & 0.9936 & \multicolumn{1}{c}{\hspace*{-8mm}1} \\[0.1cm]
		\multicolumn{1}{l|}{mMIFS-U} & 0.3964 & \multicolumn{1}{c|}{0.4024} & 0.6736 & \multicolumn{1}{c}{\hspace*{-8mm}1} \\ [0.1cm]
		\multicolumn{1}{l|}{MICC} & \multicolumn{1}{c}{\hspace*{-6mm}0} & \multicolumn{1}{c|}{0.0140} & 0.8582 & 0.9844 \\[0.1cm]
		\multicolumn{1}{l|}{QMIFS} & 0.3964 & \multicolumn{1}{c|}{0.4024} & 0.6736 & \multicolumn{1}{c}{\hspace*{-8mm}1}\\[0.1cm]
		\multicolumn{1}{l|}{NMIFS} & \multicolumn{1}{c}{\hspace*{-6mm}0} & \multicolumn{1}{c|}{\hspace*{-6mm}0}&0.2638&\multicolumn{1}{c}{\hspace*{-8mm}1}\\[0.1cm]
		\multicolumn{1}{l|}{maxMIFS} & 0.9310 & \multicolumn{1}{c|}{\hspace*{-6mm}1} & 0.9936 & \multicolumn{1}{c}{\hspace*{-6mm}1} \\
		\bottomrule
	\end{tabular}
	\label{tab:estimate_probability}
\end{table}

\begin{figure}[!h]
	\begin{minipage}[b]{0.5\linewidth}
		\centering
		\includegraphics[scale=0.35]{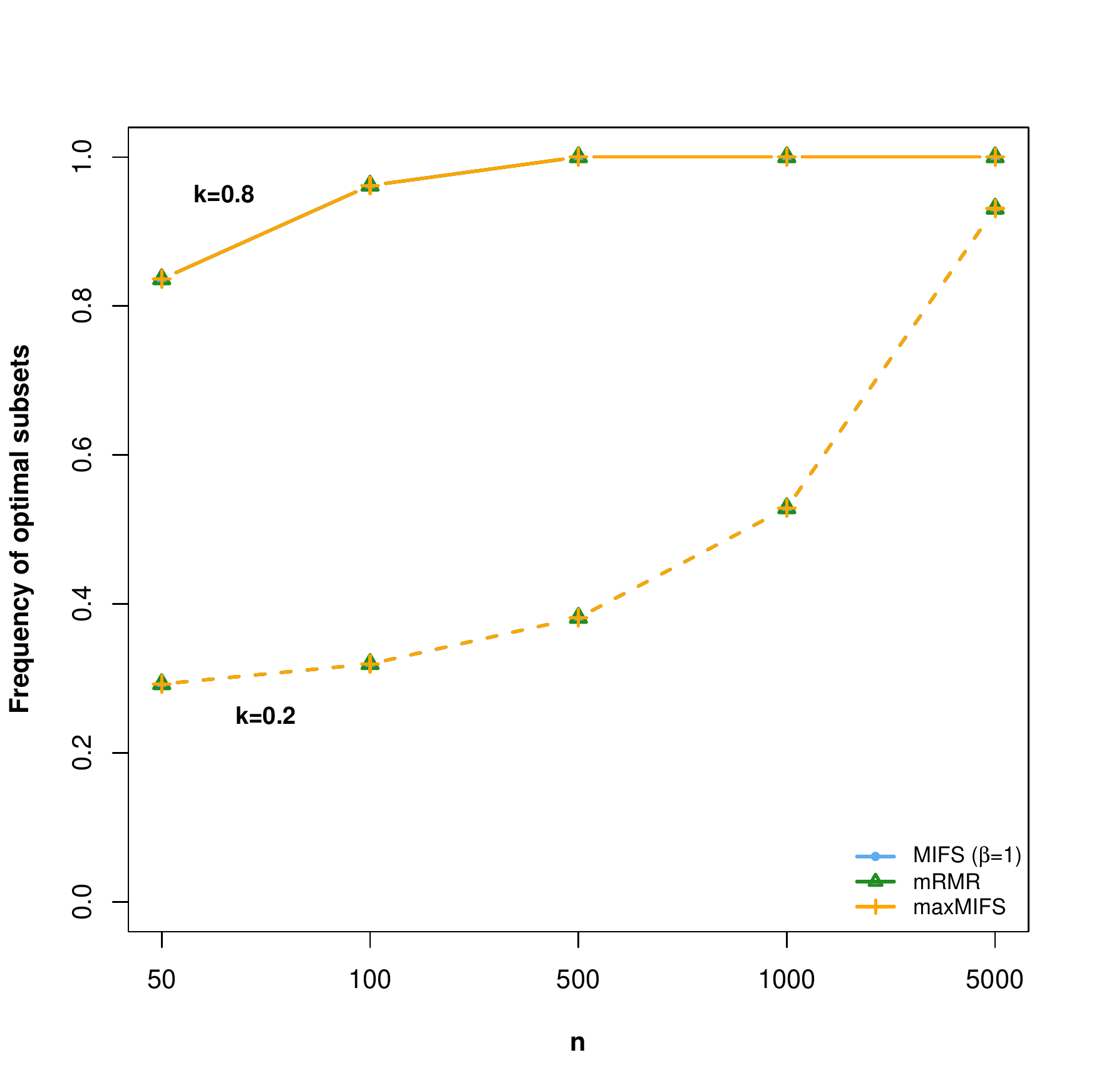}
		\begin{center}
			\vspace*{-3mm}
			(a) Scenario I\\
		\end{center}
	\end{minipage}
	\begin{minipage}[b]{0.5\linewidth}
		\centering
		\includegraphics[scale=0.35]{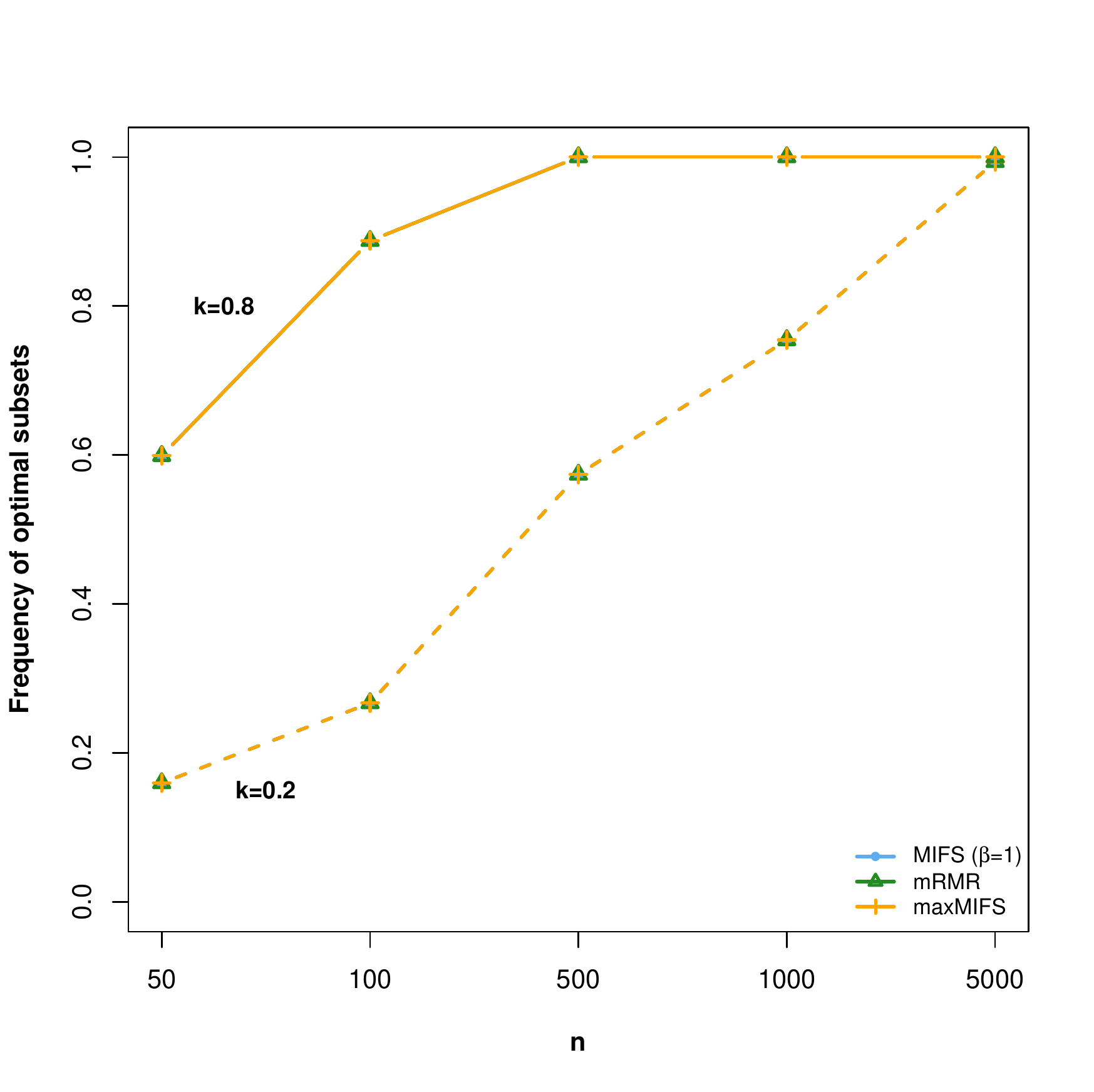}
		\begin{center}
			\vspace*{-3mm}
			(b) Scenario II\\
		\end{center}
	\end{minipage}
	\vspace*{-5mm}
	\caption{Estimated probability of selecting a relevance-optimal pair first, for MIFS ($\beta=1$), mRMR, and maxMIFS, considering $k=0.2$ and $k=0.8$, in (a) Scenario I and (b) Scenario II.}
	\label{fig:estim_relevants_scen}
\end{figure}

In Figures \ref{fig:estim_relevants_scen}(a) and \ref{fig:estim_relevants_scen}(b) we study the performance of the three best feature selection methods, MIFS ($\beta=1$), mRMR, and maxMIFS, as function of the sample size, for both scenarios and using $k=0.2$ and $k=0.8$. The performance is evaluated through the estimated probability of selecting first any of the relevance-optimal sets, i.e., $(X,Y)$, $(X,X-Y)$, $(Y,X-Y)$, $(3X+1,Y)$, or $(3X+1,X-Y)$. The results show that the performance can be significantly dependent on the sample size, especially for $k=0.2$. For $k=0.8$ the probability is close to one for $n$ equal or higher than $500$, meaning that in most cases the features are well selected. However, for $k=0.2$, good results are only obtained when $n$ is $5000$. This is due to the lack of precision in estimating the entropy and the MI. Indeed, the case $k=0.2$ is the most challenging one. Given that the strength of $Y$ is smaller in the class definition, the MI between the class and $Y$ is also smaller. For example, in Scenario I the (theoretical) MI is $0.0067$ for $k=0.2$ and $0.1153$ for $k=0.8$, as shown in Table \ref{tab:entropy_mi_true_scenario_1}. Thus, the possibility of selecting in second place features other than $Y$ is much higher. Consequently, errors in estimating the entropy and the MI, and their propagation in the calculation of the objective function, which occur with smaller sample sizes, lead to smaller performance.

In Table \ref{tab:estimate_probability} we compare the performance of the $8$ methods for the largest sample size, $n=5000$. These results confirm the poor performance of MIFS ($\beta = 0$), MIFS-U, mMIFS-U, MICC, QMIFS, and NMIFS, especially in Scenario I. There are even some cases where it was never possible to select a relevance-optimal pair of features first. This is due to unstable numerical behavior around the indeterminations of their objective functions. A probability of 0 was expected for MIFS and MIFS-U with $\beta = 0$. In these cases, $\beta=0$ removes the redundancy component, and the first two selected features are always $X$ and $3X+1$, the ones mostly associated with the class. But NMIFS also was never able to select correctly the first two features in Scenario I. In this case, the second selected feature was always $3X+1$, because the entropy estimate of $X$, despite being close to zero, was negative in all samples. Thus, the redundancy component is always negative, forcing the objective function to be maximized by the candidate feature that is most associated with already selected features, which is $3X+1$. This case illustrates the problems motivated by the sensitivity of objective functions.

\section{Conclusions}\label{sec:conclusions}

\noindent This paper undergoes an evaluation of the feature selection methods based on MI and two-dimensional sequential forward search, i.e. MIFS, MIFS-U, mRMR, mMIF-U, MICC, QMIFS, NMIFS, and maxMIFS. We clarified the differences between the MI and entropy properties of discrete and continuous random variables, whose misunderstanding has been a source of error in the proposal of feature selection methods. To support our evaluation, we established clear definitions for the notions of relevant, redundant, and irrelevant features, an issue still displaying controversy in the literature. Then, we developed a theoretical framework that allows obtaining the true feature ordering for each method under analysis, based on a scenario with two classes and a carefully chosen representative set of features. 
The ordering obtained in this way does not depend on entropy or MI estimation methods,  classifiers, or specific datasets, leading to an undoubtful comparison of the methods, which is the main contribution of our work. The feature ordering was also compared with the optimal subsets of features, obtained using the notion of relevance-optimal sets. Moreover, our framework unveiled inconsistencies in the construction of the objective functions of several feature selection methods, due to various types of indeterminations and the possibility of entropies taking null or negative values in the case of continuous random variables.

As future work, the theoretical framework established in this paper for the evaluation of two-dimensional forward feature selection methods will be extended to the case of three-dimensional methods. These methods have removed the conditional independence assumption of the two-dimensional ones and, therefore, are expected to perform better.

\section*{Acknowledgements}
This work was partially funded by Funda\c{c}\~{a}o para a Ci\^{e}ncia e a Tecnologia (FCT) through project UID/Multi/ 04621/2013. Cl\'{a}udia Pascoal also acknowledges the support of FCT via PhD grant SFRH/BD/42547/2007.



\appendix
\section*{Appendix A.}\label{Appendix}



In this appendix we prove the following theorem.

\begin{theorem}
	Let $C_k$ be defined as in (\ref{eq:class_Ck}), where $X$ and $Y$ are two independent and identically distributed features with $\textrm{Unif}\left(-\delta,\delta\right)$, $\delta >0$,  distribution. Then, $\textrm{MI}\left(C_k,X^2\right)=0$.
\end{theorem}
\noindent\textbf{Proof:}
The probability density function (pdf) of $X\sim \textrm{Unif}\left(-\delta,\delta\right)$, $\delta>0$, is
\begin{displaymath}
\dps f_{X}(x) =  \frac{1}{2\delta}\, I_{[-\delta,\delta]}(x),
\end{displaymath}
where $I_A(x)=1$ if $x\in A$ and $I_A(x)=0$ if $x\notin A$. As a consequence, the pdf of $X^2$ is
\begin{displaymath}
\dps f_{X^2}\left(u\right) = \frac{1}{2\delta\sqrt{u}}\, I_{[0,\delta^2]}(u).
\end{displaymath}

To calculate $\textrm{MI}\left(X^2,C_k\right)$ we will use definition (\ref{eq:mi_cont_disc}). Thus, we need to determine the conditional distribution of the feature $X^2$ given a class value, next derived. We start by computing the pdf of the nonnegative random variable $X^2|C_{k}=0$. As, for $u\geq 0$,
%
%
\begin{flalign*}
\dps P\left(X^2\leq u,C_k=0\right)&=
\dps \left\{
\begin{array}{c l}
\dps \int_{-\sqrt{u}}^{\sqrt{u}} \int_{-\delta}^{-\frac{x}{k}} \frac{1}{4\delta^2} \, \ud y \, \ud x &, \, 0\leq u <\left(k\delta\right)^2\\[0.2cm]
\dps \frac{k}{2}+\int_{-\sqrt{u}}^{-k\delta} \int_{-\delta}^{\delta} \frac{1}{4\delta^2} \, \ud y \, \ud x &, \, \left(k\delta\right)^2\leq u <\delta^2 \\[0.2cm]
\dps 0.5 &, \, u \geq \delta^2\\[0.2cm]
\end{array} \right. &\\[0.2cm]
&= \left\{
\begin{array}{c l}
\dps \frac{\sqrt{u}}{2\delta} &, \, 0\leq u < \delta^2\\[0.2cm]
\dps 0.5 & ,\, u \geq \delta^2\\[0.2cm]
\end{array} \right.&
\end{flalign*}
and
\begin{equation}
\label{eq:probCk01}
P(C_k=0)= P(C_k=1)=1/2.
\end{equation}
Then the pdf of $X^2|C_{k}=0$,  $f_{X^2|C_{k}=0}(\cdot)$, is given by:
\begin{equation*}
\dps f_{X^2|C_k=0}(u) = \dps \frac{1}{2\delta\sqrt{u}}\, I_{[0,\delta^2]}(u).
\end{equation*}

In sequence, for $u\geq 0$,
%
%
\begin{flalign*}
\dps P\left(X^2\leq u,C_k=1\right) &= P\left(X^2\leq u\right) - P\left(X^2\leq u,C_k=0\right) &\\
&= \begin{cases}
\frac{\dps \sqrt{u}}{\dps \delta} - \frac{\dps \sqrt{u}}{\dps 2\delta} &, \,  0\leq u < \delta^2\\
1- 0.5 &, \,  u\geq \delta^2
\end{cases}&\\
\dps  &=  \begin{cases}
\frac{\dps \sqrt{u}}{\dps 2\delta} &, \,  0\leq u < \delta^2\\
0.5 &, \,  u\geq \delta^2
\end{cases}&
\end{flalign*}
thus implying, in view of (\ref{eq:probCk01}), that $X^2|C_k=1$ has the same distribution as $X^2|C_k=0$. Therefore,
\begin{equation*}
\dps f_{X^2|C_k=1}(u) = \dps \frac{1}{2\delta\sqrt{u}}\, I_{[0,\delta^2]}(u).
\end{equation*}
Finally, we have:
%
%
\begin{flalign}
\textrm{MI}\left(X^2,C_k\right) &= 
\dps \sum_{j=0}^1 \int_{0}^{\delta^2} \hspace*{-3mm} f_{X^2|C_k=j}(u) P\left(C_k=j\right) \ln\frac{f_{X^2|C_k=j} (u)}{f_{X^2}(u)} \, \ud u
\dps&\\
& = \hspace*{-1mm} \int_{0}^{\delta^2} \hspace*{-3mm} \frac{1}{2\delta \sqrt{u}} \frac{1}{2} \ln\left(\frac{\frac{1}{2\delta \sqrt{u}}}{\frac{1}{2\delta \sqrt{u}}}\right) \,\ud u = 0.&
\end{flalign}
\begin{flushright} $\square$\end{flushright}
We should stress that the result of the theorem, namely $\textrm{MI}\left(X^2,C_k\right)=0$, still holds with the $\textrm{Unif}\left(-\delta,\delta\right)$ distribution substituted by an arbitrary symmetric absolutely continuous distribution.

\section*{References}

\bibliographystyle{wileyj}  
\bibliography{reftese}

\end{document}